%% file: main.tex
\pdfoutput=1
\documentclass[10pt,twocolumn,letterpaper]{article}

\usepackage{iccv}
\usepackage{times}
\usepackage{epsfig}
\usepackage{graphicx}
\usepackage{amsmath}
\usepackage{amssymb}

\usepackage{caption} 
\usepackage{titling}
 \usepackage{xspace}
\usepackage{algorithm}
\usepackage{algorithmic}
\usepackage{listings}
\usepackage{amsmath}
\usepackage{amsthm, amssymb}
\usepackage{pifont}%
\usepackage{bbm}
\usepackage{array}
\usepackage{wrapfig}
\usepackage{multirow}
\usepackage{tabularx}
\usepackage[table]{xcolor}
\usepackage{booktabs}
\usepackage{makecell}
\usepackage{cellspace}
\usepackage{stackengine}
\usepackage{bm}
\usepackage{mathtools}
\usepackage[switch]{lineno}
\newcommand{\xmark}{\ding{55}}%
\newcommand{\band}{\rowcolor{gray!10}}

\newcommand\Mycomb[2][^n]{\prescript{#1\mkern-0.5mu}{}C_{#2}}
\setstackEOL{\\}
\usepackage{etoolbox}
\makeatletter
\AfterEndEnvironment{algorithm}{\let\@algcomment\relax}
\AtEndEnvironment{algorithm}{\kern2pt\hrule\relax\vskip3pt\@algcomment}
\let\@algcomment\relax
\newcommand\algcomment[1]{\def\@algcomment{\footnotesize#1}}
\renewcommand\fs@ruled{\def\@fs@cfont{\bfseries}\let\@fs@capt\floatc@ruled
  \def\@fs@pre{\hrule height.8pt depth0pt \kern2pt}%
  \def\@fs@post{}%
  \def\@fs@mid{\kern2pt\hrule\kern2pt}%
  \let\@fs@iftopcapt\iftrue}
\makeatother
\input{utils/macros.tex}

\usepackage[pagebackref=true,breaklinks=true,letterpaper=true,colorlinks,bookmarks=false]{hyperref}

\iccvfinalcopy % *** Uncomment this line for the final submission

\def\iccvPaperID{6026} % *** Enter the ICCV Paper ID here

\ificcvfinal\fi

\begin{document}

%%%%%%%%% TITLE
\title{\textbf{Contrast and Classify: Training Robust VQA Models}}

\author{
    Yash Kant$^{1}$\thanks{Correspondence to \texttt{ysh.kant@gmail.com}}\quad
    Abhinav Moudgil$^{1}$\quad
    Dhruv Batra$^{1,2}$\quad
    Devi Parikh$^{1,2}$\quad
    Harsh Agrawal$^{1}$\quad \\
    $^1$Georgia Institute of Technology\quad
    $^2$
    Facebook AI Research\\
}
\date{}
\maketitle
% Remove page # from the first page of camera-ready.
\ificcvfinal\thispagestyle{empty}\fi

\input{sections/abstract}
\input{sections/intro-iccv.tex}

\input{utils/space_saver.tex}
\input{sections/related}

\input{sections/method_final}

\input{sections/experiments}
\input{sections/results}

{\small
\bibliographystyle{ieee_fullname}
\bibliography{references}
}

\clearpage
%%%%%%%%% TITLE
\title{\textbf{Appendix}}
\author{}
\date{}
\maketitle

\setcounter{section}{0}
\def\thesection{\Alph{section}}
\input{supp-sections/supp_experiments}

\end{document}

%% file: utils/macros.tex
\newcommand{\Mod}[1]{\ (\mathrm{mod}\ #1)}

\makeatletter
\DeclareRobustCommand\onedot{\futurelet\@let@token\@onedot}
\def\@onedot{\ifx\@let@token.\else.\null\fi\xspace}
 
\def\ie{\emph{i.e}\onedot}

\def\etal{\emph{et al}\onedot}
\makeatother

\def\bt/{back-translation} % use \bt/
\def\BT/{Back-translation} % use \bt/
\def\vqg/{visual question generation} % use \vqg/ and so on
\def\VQG/{VQG} % use \vqg/ and so on
\def\SCL/{SCL} % use \vqg/ and so on
\def\uni/{UNITER} % use \vqg/ and so on
\def\CE/{Cross-entropy}
\def\ce/{cross-entropy}
\def\concat/{ConClaT}
\newcommand\lscl[1]{$\mathcal{L}_{\text{SC}}$}
\newcommand\lsscl[1]{$\mathcal{L}_{\text{SSC}}$}
\newcommand\lce[1]{$\mathcal{L}_{\text{CE}}$}
\newcommand\lnce[1]{$\mathcal{L}_{\text{NCE}}$}

\def\xpos{\mathcal{X}^{+}}
\def\xneg{\mathcal{X}^{-}}
\def\xpara{\mathcal{X}^{+}_{\texttt{para}}}
\def\xcls{\mathcal{X}^{+}_{\texttt{cls}}}

\def\lssc{\mathcal{L}_{\text{SSC}}}

%% file: sections/abstract.tex
\begin{abstract}

Recent Visual Question Answering (VQA) models have shown impressive performance on the VQA benchmark but remain sensitive to small linguistic variations in input questions. Existing approaches address this by augmenting the dataset with question paraphrases from visual question generation models or adversarial perturbations. These approaches use the combined data to learn an answer classifier by minimizing the standard \ce/ loss. To more effectively leverage augmented data, we build on the recent success in contrastive learning. We propose a novel training paradigm (\concat/) that optimizes both \ce/ and contrastive losses. The contrastive loss encourages representations to be robust to linguistic variations in questions while the \ce/ loss preserves the discriminative power of representations for answer prediction. 

We find that optimizing both losses -- either alternately or jointly -- is key to effective training. On the VQA-Rephrasings~\cite{shah2019cycle} benchmark, which measures the VQA model's answer consistency across human paraphrases of a question, \concat/ improves Consensus Score by 1.63\% over an improved baseline. In addition, on the standard VQA 2.0 benchmark, we improve the VQA accuracy by 0.78\% overall. We also show that \concat/ is agnostic to the type of data-augmentation strategy used.  

\end{abstract}
\vspace{-10pt}

% We extensively ablate \concat/ with various numerous training strategies~\cite{chen2019complement, Khosla2020SupervisedCL}, and compare with previously proposed losses~\cite{free_vqa, patro2018differential} to justify our choices. 

% and training VQA models by curating batches with  our proposed negative sampling strategy further boosts performance. 
% On the VQA Rephrasings benchmark which measures the model's answer consistency across several rephrasings of a question, \concat/ improves Consensus Score~\cite{shah2019cycle} by 1.63\% over an improved baseline. In addition, on the standard VQA 2.0 benchmark, we improve the VQA accuracy by 0.78\% overall. It is also worth noting that under our training paradigm (\concat/), VQA models show better performance than existing approaches across both the aforementioned data-augmentation strategies -- Back Translation and VQG.

% VQA models trained with ConClaT achieve higher consensus scores on the VQA-Rephrasings dataset as well as higher VQA accuracy on the VQA 2.0 dataset compared to existing approaches across a variety of data augmentation strategies. 

%% file: sections/intro-iccv.tex
\section{Introduction}

\begin{figure}[t!]
\centering
\includegraphics[width=0.9\linewidth]{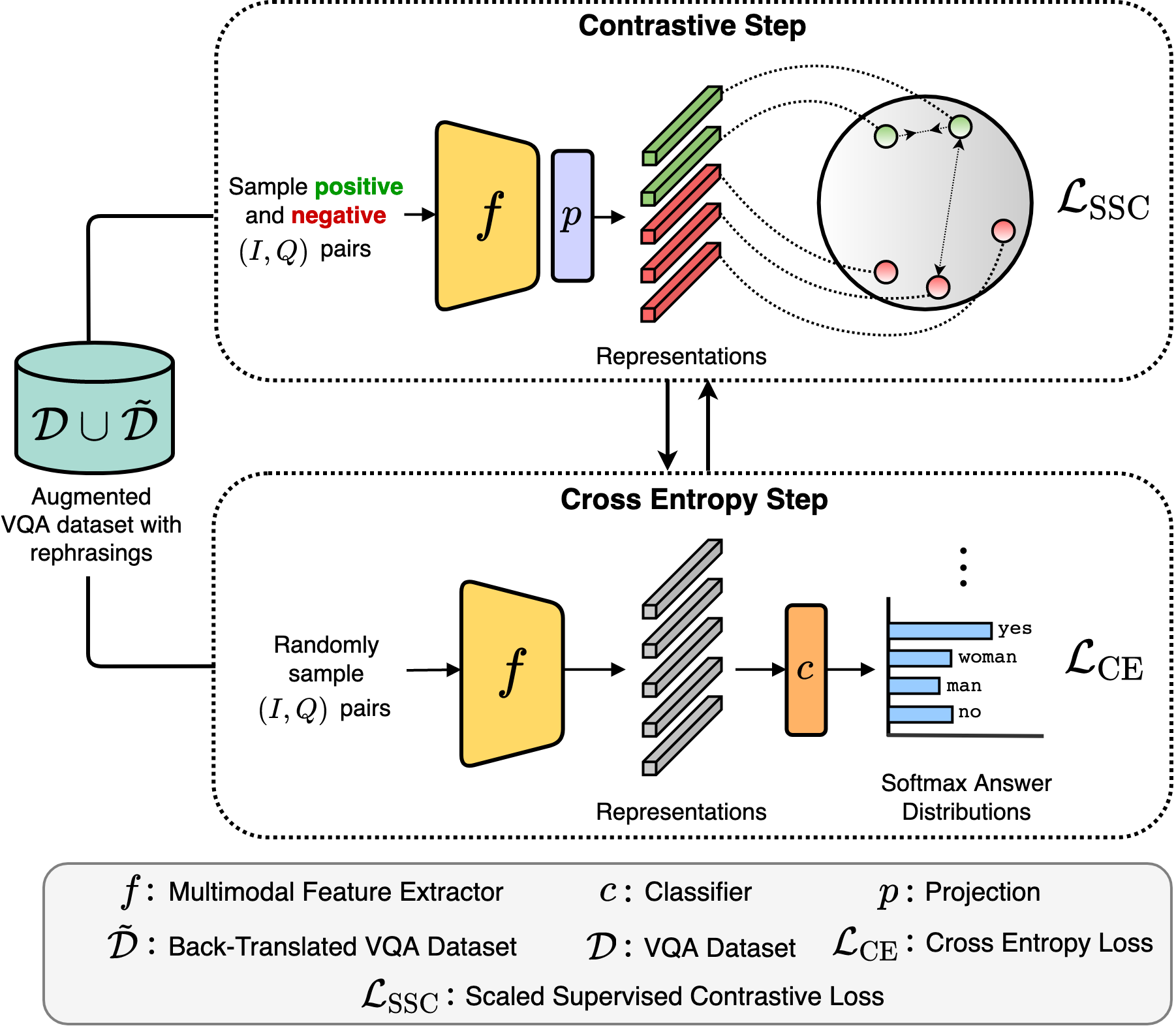}
\caption{We make VQA model robust to question paraphrases using a training paradigm ConClaT that minimizes contrastive and \ce/ losses together. Contrastive learning step pulls representations of positive samples corresponding to paraphrased questions closer together while pushing those with different answers farther apart. \CE/ step makes these representations discriminative to help model answer visual questions accurately.}
\label{fig:teaser}
\vspace{-15pt}
\end{figure}

% \item VQA models are inconsistent and brittle

% \item Explain what is inconsistency and why it matters. [reliablity etc.]

Visual Question Answering (VQA) refers to the task of automatically answering free-form natural language questions about an image. For VQA systems to work reliably when deployed in the wild, for applications such as assisting visually impaired users, they need to be robust to different ways a user might ask the same question. 
For example, VQA models should produce the same answer for two paraphrased questions – ``What is in the basket?'' and ``What is contained in the basket?'' since their semantic meaning is the same. While significant progress has been made towards building more accurate VQA systems, these models remain brittle to minor linguistic variations in the input question.

% VQA model <Q1> vs <parphrased Q1> result in two entirely different answers whereas to us the both mean the same thing. 
% In order to benchmark progress of VQA models previous work \cite{} collected human rephrasings on the subset original VQA 2.0 validation set. 

% \item Explain how present-day VQA models learn joint V+L representations
% \item Explain how existing approaches tackle robustness in VQA

To make VQA systems robust, % to small linguistic variations, 
existing approaches~\cite{shah2019cycle,tang2020semantic} have trained VQA systems~\cite{jiang2018pythia} by augmenting the training data with different variations of the input question. For instance, VQA-CC~\cite{shah2019cycle} use a visual question generation (VQG) model to generate paraphrased question given an image and answer. %where as ~\cite{tang2020semantic} uses back-translation to produce question rephrasings. 
Generally, these models fuse image and question features into a joint vision and language (V+L) representation followed by a standard softmax classifier to produce answer probabilities and are optimized by minimizing the \ce/ loss. Unfortunately, \ce/ loss treats every image-question pair independently and fails to exploit the information that some questions in the augmented dataset are paraphrases of each other.

We overcome this limitation by using a contrastive loss InfoNCE~\cite{Oord2018RepresentationLW} that encourages joint V+L (Vision and Language) representations obtained from samples whose questions are paraphrases of each other to be closer while pulling apart the V+L representations of samples with different answers. As we operate in a supervised setting, we choose Supervised Contrastive Loss (\SCL/)~\cite{Khosla2020SupervisedCL} which extends InfoNCE by utilizing the label information to bring samples from the same class (ground-truth answer) together. We introduce a variant of the \SCL/ which emphasizes rephrased image-question pairs over pairs that are entirely different but have the same answer.
% \ha{We introduce a variant of the \SCL/ loss (\lsscl/) which assigns higher weight to augmented positive samples (rephrasings in our case) over intra-class positive samples (that have the same answer).} 
Our proposed training paradigm, \concat/ (\textbf{Con}trast and \textbf{Cla}ssify \textbf{T}raining), minimizes \SCL/ and \ce/ loss together to learn better vision and language representations as shown in Fig.\ref{fig:teaser}. Minimizing the contrastive loss encourages representations to be robust to linguistic variations in questions while the \ce/ loss preserves the discriminative power of the representations for answer classification. Instead of pretraining with \SCL/, then fine-tuning with \ce/ loss as in~\cite{Khosla2020SupervisedCL}, we find that minimizing the two losses either alternately or jointly by constructing loss-specific mini-batches helps learn better representations. 
% Inspired by the use of hard negatives in classification tasks~\cite{1467360, 5255236, 6126229}, we define various types of negatives and positives given a reference \yk{sample}\ha{reference what?}. We also propose a batch curation strategy to construct mini-batches by sampling from these negatives and positives for the contrastive loss.
For contrastive loss, we carefully curate mini-batches by sampling various types of negatives and positives given a reference sample.

We show the efficacy of our training paradigm across two rephrasing (i.e., data-augmentation)  strategies. Using rephrasings obtained from a VQG model proposed in~\cite{shah2019cycle}, our approach outperforms 
% \ha{on what?}
a baseline that simply treats these rephrasings as additional samples and ignores the link between question and its paraphrases. We noticed that the VQG model fails to produce a diverse set of rephrasings for a question. Hence, we use \BT/ to obtain question rephrasings. \BT/ \cite{edunov2018understanding} involves translating an input sentence from one language to another and then translating it back into the original language using a pair of machine translation models (e.g. \texttt{en-fr} and \texttt{fr-en}). We find that \BT/ preserves the semantic meaning of the question while generating syntactically diverse question. Utilizing the publicly available collection of neural machine translation models in HuggingFace~\cite{Wolf2019HuggingFacesTS}, we generate numerous rephrasings of every question. Then, we filter poor/irrelevant rephrasings with a sentence similarity model \cite{reimers2019sentencebert} and store 3 rephrasings per original question of VQA v2.0 dataset without any manual supervision.

We extensively ablate \concat/ with alternate~\cite{chen2019complement}, joint and pretrain-finetune~\cite{Khosla2020SupervisedCL} training schemes, and compare with previously proposed triplet~\cite{patro2018differential} and margin-based losses~\cite{free_vqa} 
% \ha{can we briefly mention what these strategies are and what the losses are? Otherwise the reader will have to go search the paper. If these are well known losses then just name them before the citation.} 
. We evaluate on the VQA Rephrasings benchmark~\cite{shah2019cycle} which measures the model's answer consistency across several rephrasings of a question. \concat/ improves Consensus Score by 1.63\% over an improved baseline. In addition, on the standard VQA 2.0 benchmark, we improve VQA accuracy by 0.78\% overall. It is also worth noting that VQA models trained using \concat/ perform better than existing approaches across both the aforementioned data-augmentation strategies -- \BT/ and VQG.

% % \item Summarize Contributions [-Alternate \SCL/ -Hard Mining -BT data?]

% % Overall we improve --- on VQA-Reprasings and --- on VQA 2.0 validation set. Our contributions are as follows: --- \yk{Expand more}
% To summarize, we make the following contributions.
% \begin{itemize}
%     \item We propose a novel training paradigm that involves alternate training of contrastive and \CE/ losses to learn joint vision and language representations that are robust to linguistic variations in input questions. 
%     \item We automatically collect a large-scale dataset of X paraphras questions to complement the original VQA dataset using back translation with publically available neural machine translation models. 
%     \item We emperically show that our proposed approach improves consistency score and VQA scores on the VQA-Rephrasings and VQA2.0 dataset respectively across a variety of data augmentation techniques. 
% \end{itemize}

%% file: utils/space_saver.tex
% Some illegal space-saving macros
\captionsetup[table]{skip=7pt}
\setlength{\textfloatsep}{3pt}% Remove \textfloatsep

%% file: sections/related.tex
\section{Related Work}
% \begin{itemize}
% \item VQA models\\
% - Basic fusion approaches that fuse CNN grid features and LSTM representations using different forms of attention~\cite{hiecoatt,san,fukui2016multimodal,Jiang2020InDO}\\
% - Approaches that use object detector features~\cite{bua,kim2018bilinear,yu2018beyond,jiang2018pythia}\\
% - More recently, multi-modal transformers have gained popularity. Features from object detector and BERT language transformer 
% are fed as input a transformer architecture. The attended representations are used to learn a 1000-way classifier on top.~\cite{lu2019vilbert,lu202012,su2019vl,li2020unicoder,su2019vl,tan2019lxmert,chen2019uniter}.\\
% - We use the latest state-of-the-art multi-modal transformer based architectures as our model for all the experiments.\\
\textbf{Models for VQA.} 
Several models have been proposed for Visual Question Answering which fuse CNN grid features and LSTM features with different forms of attention~\cite{hiecoatt,san,fukui2016multimodal,Jiang2020InDO}. Bottom-Up and Top-Down~\cite{bua} proposed to learn attention over object regions obtained from a pretrained object detector and subsequent works~\cite{kim2018bilinear,yu2018beyond,jiang2018pythia} introduced various ways to fuse image and language representations. Recent works~\cite{lu2019vilbert,lu202012,su2019vl,li2020unicoder,su2019vl,tan2019lxmert,chen2019uniter} use multi-modal transformers to learn visuo-linguistic representations from object detector features and BERT question features~\cite{devlin2018bert}. We use the multi-modal transformer architecture similar to UNITER~\cite{chen2019uniter} for all our experiments.

% \item Robustness of VQA models. Consistency works. \\
% - Robustness Dataset Bias~\cite{goyal2016making,zhang2015yin} and language priors~\cite{gvqa,cvqa,ramakrishnan2018overcoming}\\
% - Consistency between answers to questions and sub-questions.~\cite{selvaraju2020squinting}\\
% - Robustness to perturbations in images and questions~\cite{zeng2019adversarial,tang2020semantic}\\
% - Robustness to paraphrases of questions~\cite{shah2019cycle}. We specifically look at this. \\
% Several efforts have been made to study and address the robustness of VQA models with respect to multi-modal vision and language input
\textbf{Robustness of VQA Models.}
Robustness of VQA models with respect to multi-modal vision and language input has been studied in great detail. \cite{goyal2016making,zhang2015yin} proposed balanced datasets to ensure models don't overfit to language while answering visual questions. C-VQA~\cite{cvqa} and VQA-CP~\cite{gvqa} datasets were proposed to test robustness against changing question-answer distributions. SQUINT~\cite{selvaraju2020squinting} encouraged consistency between reasoning questions and associated sub-questions. Our work focuses on robustness to question paraphrases in  VQA-Rephrasings~\cite{shah2019cycle} that were collected from human annotators. VQA-CC~\cite{shah2019cycle} trained a Visual Question Generation (VQG) model to generate paraphrases of questions to augment the training dataset while VQA-Aug~\cite{tang2020semantic} augmented the training dataset by generating paraphrases of questions via \bt/. We show that these data augmentation techniques can be better utilized via \concat/ to build robust and accurate VQA models. Concurrent to our work, Whitehead \etal\cite{whitehead2020learning} propose a rule-based mechanism to generate question paraphrases for VQA. They constrain their model architecture to be modular~\cite{nmn} and use module-level loss to improve consistency. In contrast, our approach is agnostic to model architecture.

% Orthogonal to robustness to minor linguistic variations, various works~\cite{fu2020counterfactual, chen2020counterfactual, agarwal2020causal, teney2020learning} have also studied methods to improve counterfactual answering in VQA. 
% Various works~\cite{} made the VQA models robust to language bias (For e.g, ``What is the color of x'' will always produce `blue' irrespective of x)
% VQA models have shown a tendency to be biased towards language, For e.g..., various works~\cite{} have made the VQA models robust to it. 

% Various works ~\cite{fu2020counterfactual, chen2020counterfactual, agarwal2020causal, teney2020learning} made the VQA models robust to language bias (For e.g, ``What is the color of x'' will always produce `blue' irrespective of x) and improved counterfactual answering (answer should change when the semantic content of the question or the image changes). Our work focuses on robustness to syntactic variations in questions as opposed to semantic variations. 
Various works~\cite{gvqa, agarwal2020causal, free_vqa, patro2018differential} made VQA models robust to language bias (For example, “What is the color of $x$” will always produce ‘blue’ irrespective of $x$). Recent works~\cite{teney2020learning, abb2020conterfactual, chen2020counterfactual, pan2019questionconditioned} also studied robustness from counterfactual answering lens – answer should change according to the change in semantic content of the question or image.  Our work, on the other hand, focuses on robustness to \textit{syntactic} variations in questions.

% \ha{Aren't we missing discussing all the papers that were mentioned in the reviews?}

% \item Rephrase Generation \\
% - Visual Question Generation~\cite{shah2019cycle}\\
% - Large body of work on paraphrase generation in NLP using LSTM networks ~\cite{prakash2016neural}, deep reinforcement learning~\cite{li2017paraphrase}, transformer models~\cite{wang2019task}
% and VAE based architectures~\cite{gupta2017deep} but they require supervision in the 
% form of paraphrase pairs. \\
% - To generate paraphrases in a self-supervised manner, 
% neural machine translation has been used to generate rephrasings in NLP~\cite{mallinson2017paraphrasing,wieting2017learning}. We build on top 
% of these works and use the latest state-of-the-art NMT models from HuggingFace~\cite{} to generate paraphrases for visual questions 
% without any paired supervision. \\
\textbf{Paraphrase Generation in NLP.}
There has been significant work in the area of Natural Language Processing (NLP) for generating paraphrases of a sentence using LSTM networks~\cite{prakash2016neural}, Deep Reinforcement Learning~\cite{li2017paraphrase}, Variational Autoencoders~\cite{gupta2017deep} and Transformers~\cite{wang2019task}. However, these works require supervision in the form of paraphrase pairs. In order to mitigate this limitation of labelled data, Neural Machine Translation (NMT) models have been used to generate paraphrases in a self-supervised fashion via \bt/~\cite{mallinson2017paraphrasing,wieting2017learning}. We build on top of these works and use state-of-the-art NMT models from HuggingFace~\cite{Wolf2019HuggingFacesTS} to generate paraphrases for visual questions without any supervision.
\textbf{Contrastive Learning.} There has been recent interest in the use of Contrastive Learning for learning visual representations in a self-supervised
manner~\cite{wu2018unsupervised,henaff2019data,he2020momentum,chen2020simple,chen2020improved,chen2020big,radford2021learning}. 
Going beyond Image Classification, recently, ~\cite{gupta2020contrastive} used contrastive learning for phrase grounding. They used the InfoNCE loss~\cite{Oord2018RepresentationLW} to learn a compatibility function between a set of region features from an image and contextualized word representations. In contrast, we want to learn representations which are robust to linguistic variations in the question for VQA. 

To utilize label information in contrastive losses,~\cite{Khosla2020SupervisedCL} proposed Supervised Contrastive Learning (SCL) loss for learning \textit{visual} representations. We introduce a variant of the SCL which scales the contributions from augmented positive samples (rephrasings in our case) over intra-class positive samples (that have the same answer) using a scaling factor. Morever, our training paradigm optimizes both (\ce/ and \SCL/) losses together, whereas~\cite{Khosla2020SupervisedCL} follow the pretrain-finetune training scheme. Furthermore,~\cite{Khosla2020SupervisedCL} randomly sample positive and negative pairs based on label information, whereas we carefully curate batches by sampling hard-negatives from the dataset. We show how these differences affect performance through a series of ablations in our experiments section.

%% file: sections/method_final.tex
\begin{figure*}[t!]
  \centering
  \includegraphics[width=0.87\textwidth]{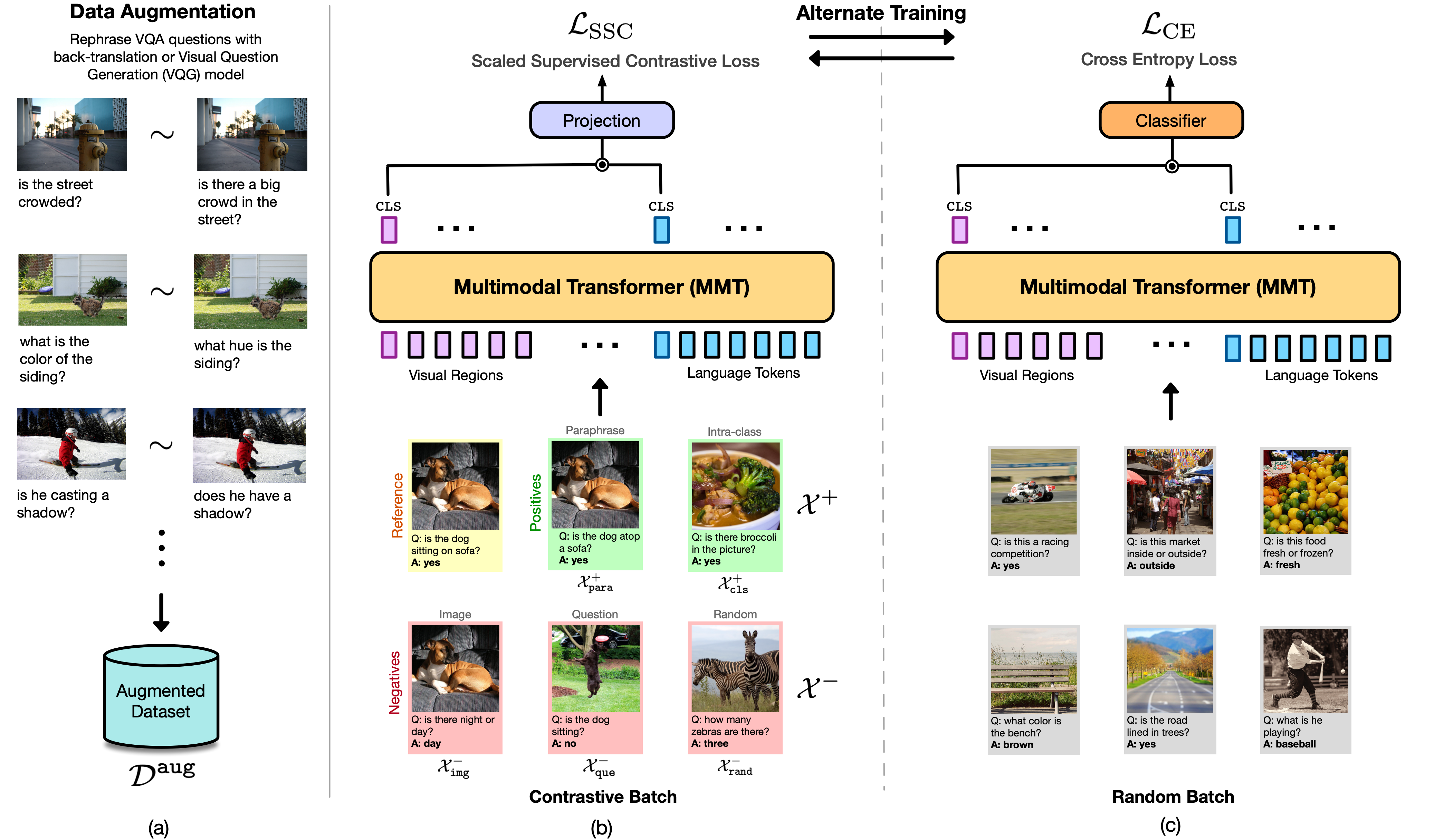}
  \caption{\textbf{Overview of ConClaT.} (a) We augment the VQA dataset by paraphrasing every question via \BT/ or Visual Question Generation. 
  %Back-translation preserves semantic meaning of the question but generates diverse set of paraphrases for each question. 
  (b) We carefully curate a contrastive batch by sampling different types of positives and negatives to learn joint V+L representations by minimizing scaled supervised contrastive loss \lsscl/. (c) Cross Entropy loss \lce/ is optimized with \lsscl/.}
  \label{fig:approach}
  \vspace{-10pt}
\end{figure*}

\section{Preliminaries}
% In this section, we first introduce notations and expand on our VQA architecture that jointly learns V+L representations. 
% Next, we present our modified formulation of Supervised Contrastive Loss \cite{Khosla2020SupervisedCL} 
% for our use case. Finally, expand on the details of our alternate training paradigm and mini-batch creation
% process for the contrastive loss. 
% In this section, we introduce the VQA task and the typical cross entropy training of the VQA models. We then recap contrastive learning and the recently proposed Supervised Contrastive Learning (\SCL/) \cite{Khosla2020SupervisedCL} loss.
In this section, we introduce the VQA task and the standard cross entropy training of VQA models. We then recap contrastive methods for learning representations~\cite{chen2020simple} and the recently proposed Supervised Contrastive Learning (\SCL/) \cite{Khosla2020SupervisedCL} setup. We describe our approach in section~\ref{sec:approach}. \smallskip
%discuss our proposed training paradigm to train VQA models which are robust to minor linguistic variations in the questions. We start with an overview of the VQA dataset and how to augment the dataset with question rephrasings. We then introduce our training paradigm that involves alternately optimizing cross-entropy loss with our variant of the Supervised Contrastive Learning (SCL)~\cite{Khosla2020SupervisedCL} loss. Finally, we describe how carefully curating batches for the two losses can help learn more robust VQA models. 

\noindent \textbf{VQA.} The task of Visual Question Answering (VQA)~\cite{vqa,goyal2016making} involves predicting an answer $a$ for a question $q$ about an image $v$. An instance of this problem in the VQA Dataset $\mathcal{D}$ is represented via a tuple $x=(v, q, a),\forall x \in \mathcal{D}$. Recent VQA models~\cite{jiang2018pythia,bua,chen2019uniter} take
image and question as input and output a joint vision and language (V+L) representation $\bm h \in \mathcal{R}^{d_h}$ using a multi-modal network $f$:
% $$\bm{h} = f^{mmt}(v, q; \theta^{mmt})$$
$$\bm{h} = f(v, q)$$

The V+L representation $\bm{h}$ is then used to predict a probability distribution over the answer space $\mathcal{A}$ with a softmax classifier $f^{c}(\bm h)$ learned by minimizing the cross-entropy:

\begin{equation}
\begin{aligned}
        \mathcal{L}_{\text{CE}} = -\text{log}\frac{\exp(f^c(\bm h)[a])}{\sum_{a'\in \mathcal{A}}\exp(f^c(\bm h)[a'])}
        % \bm h &= f^{mmt}(v,q)\\
        % \mathcal{L}_{ce} = \min_{\theta} &-\mathbb{E_{\mathcal{X}}}[\log p_{\theta}(a|v,q)]
\end{aligned}
\end{equation}
where $f^c(\bm h)[a]$ is the logit corresponding to the answer $a$. \smallskip

\noindent \textbf{Contrastive Learning.} Recent works in vision \cite{chen2020simple} have used contrastive losses to bring representations of two augmented views of the same image (called positives) closer together while pulling apart the representations of two different images (called negatives). %Following recent works in contrastive learning \cite{chen2020simple,Khosla2020SupervisedCL}, we want to bring joint V+L representations obtained from  the question and its paraphrase to be closer together while pulling apart the V+L representations for questions with different answers. 
The representation $\bm h$ 
%using a Convoluational Neural Network \cite{krizhevsky2012imagenet} 
obtained from an image encoder is projected into a $d_z$ dimensional hyper-sphere
%normalized vector 
using a projection network $g$ such that $\bm z = g(\bm h) \in \mathcal{R}^{d_z}$. Given a mini-batch of size $K$, the image representation $\bm h$ is learned by minimizing the InfoNCE~\cite{Oord2018RepresentationLW} loss which operates on a pair of positives $(\bm z_i,\bm z_p)$ and $K-1$ negative pairs $(\bm z_i,\bm z_k)$ such that $ i,p,k \in [1, K], k\neq i$ as follows:

\begin{equation}
\label{eq:nce_loss}
    \mathcal{L}_{\text{NCE}}^{i} = -\log \frac{\exp(\Phi(\bm z_i, \bm z_p)/\tau)}{\sum_{k=1}^{K} \mathbbm{1}_{k \neq i}\exp(\Phi(\bm z_i, \bm z_k)/\tau)}~,
\end{equation}
where $\mathrm{\Phi}(\bm u,\bm v) = \bm u^\top \bm v / \lVert\bm u\rVert \lVert\bm v\rVert$ computes similarity between $\bm u$ and $\bm v$ and $\tau > 0$ is a scalar temperature parameter.

A generalization of InfoNCE loss to handle more than one positive-pair was proposed by \cite{Khosla2020SupervisedCL} called Supervised Contrastive Loss (SCL). Given a reference sample $x$, SCL uses class-label information to form a set of positives $\mathcal{X}^{+}(x)$ that contains samples with the same label as $x$. $\mathcal{X}^{+}(x)$ also contains augmented views of the sample because they share the same label as $x$. For a minibatch with $K$ samples, \SCL/ is defined as:
% The \SCL/ loss given a reference sample $x_i$ and corresponding set of positives denoted by $\mathcal{P}(x_i)$ 
% The set of positives $\mathcal{P}(x_i)$ also contain augmented views of the sample $x_i$ since they also have the same ground-truth label as $x$.
\begin{gather}
%\begin{align}
\mathcal{L}_{\text{SC}}^{i} = -\sum_{p=1}^{\vert \mathcal{X}^{+}(x_i) \vert} { \log \frac{\exp(\Phi(\bm z_i.\bm z_p)/\tau)}{\sum_{k=1}^{K}{\mathbbm{1}_{k \neq i}
\cdot \exp(\Phi(\bm z_i.\bm z_p)/\tau)}}}\nonumber \\
\mathcal{L}_{\text{SC}} = \sum_{i = 1}^{K}{\frac{\mathcal{L}_{\text{SC}}^{i}}{\vert \mathcal{X}^{+}(x_i) \vert}} \label{eq:scl_i}
%\end{align}
 \end{gather}
Overall, $\mathcal{L}_{\text{SC}}^{i}$ tries to bring the representation of samples in $ \mathcal{X}^{+}(x_i)$ closer together compared to representations of samples with a different ground-truth label.

\section{Approach} \label{sec:approach}
We now describe our approach, \concat/, which uses contrastive and cross-entropy training to learn VQA models robust to question paraphrases.

\noindent \subsection{Augmented Dataset with \BT/}  We augment the train set with question paraphrases using 88 different MarianNMT~\cite{mariannmt} \BT/ model pairs released by HuggingFace~\cite{Wolf2019HuggingFacesTS}. We produce ~ 27 \emph{unique} rephrasings per question with cosine similarity of 0.88 on average, the similarity is calculated by first encoding the questions via Sentence-BERT \cite{reimers2019sentencebert}. We only select paraphrases that have $\geq 0.95$ similarity with the original question and choose three unique paraphrases randomly from this subset. We use three paraphrases to keep the compute manageable. Overall, our augmented train set consists of $\sim$ 1.6M samples.

For a sample $x = (v, q, a) \ \in \mathcal{D}$, let's denote a set of paraphrases for question $q$ by $\mathcal{Q}(q)$ and the corresponding 
set of VQA triplets as: 

\begin{equation}
    \xpara(x) = \{(v, q', a) \mid  q' \in \mathcal{Q}(q) \}
\end{equation} 
% $$$$

As shown in Figure~\ref{fig:approach}(a), we augment the VQA dataset $\mathcal{D}$ with multiple paraphrased samples of a given question and denote the augmented dataset $\mathcal{D}^\texttt{aug}$ as:

\begin{equation}
    \mathcal{D}^{\texttt{aug}} = \mathcal{D} \bigcup_{x \in \mathcal{D}}\xpara(x)    
\end{equation} 

\noindent \subsection{Scaled Contrastive Loss for VQA} We would like our VQA model to produce the \textit{same and correct} answer for a question and its paraphrase given an input image. This motivates us to map joint vision and language (V+L) representations of an original and paraphrased sample closer to each other. Moreover, since we operate in a supervised setting, following SCL~\cite{Khosla2020SupervisedCL} we also pull the joint representations for the questions with the same answer (intra-class positives) closer together while pulling apart the representations of questions with different answers. We define the set of all samples with the same ground truth answer as $x$ by: 
% $$\xpos(x) = \{ \hat{x} = (\hat{v}, \hat{q}, a) \mid \hat{x} \in \mathcal{D}^{aug} \}$$
%$$\xpos(x) = \{ (\hat{v}, \hat{q}, \hat{a}) \mid \hat{a} = a, ~\forall~(\hat{v}, \hat{q}, \hat{a}) \in \mathcal{D}^{aug}  \}$$

\begin{equation}
    \xpos(x) = \{ (\hat{v}, \hat{q}, \hat{a}) \in \mathcal{D}^{\texttt{aug}} \mid \hat{a} = a \}
\end{equation}

Note that $\xpara(x) \subset \xpos(x)$ as all question paraphrases have the same answer for a given image but not all questions with the same answer are paraphrases. We refer to samples in set $\xcls(x) = \xpos(x) - \xpara(x)$ as \textit{intra-class} positives and set $\xpara(x)$ as \textit{paraphrased} positives w.r.t. $x$ as depicted in Figure~\ref{fig:approach}(b).  
% We generate these paraphrases either by using \BT/~\cite{edunov2018understanding} or a visual question generation (\VQG/) model~\cite{shah2019cycle} .

Following Eq.~\eqref{eq:scl_i}, all the samples in $\xpos(x_i)$ in \lscl/ are treated the same. That is, representations from both the paraphrased positives and intra-class positives are brought closer together. 
% Similar to using just cross-entropy loss on the augmented dataset, 
% \yk{Motivate this by saying we do this for Robust VQA models}
To emphasize on the link between question and its paraphrase, 
% SCL loss fails to emphasize the link between question and its paraphrase, to overcome this, 
we propose a variant of the SCL in Eq.~\eqref{eq:sscl_i} which assigns higher weight to paraphrased positives $\xpara(x)$ over intra-class positives $\xcls(x)$. 
We introduce a scaling factor $\alpha_{ip}$ in the SCL (Eq.~\eqref{eq:scl_i}) for a sample $x_i$ as follows: 
\begin{equation}
\lssc^{i} = -\sum_{p=1}^{\vert \xpos(x_i) \vert} {\alpha_{ip} \cdot \log \frac{\exp(\Phi(\bm z_i.\bm z_p)/\tau)}{\sum_{k=1}^{K}{\mathbbm{1}_{k \neq i}
\cdot \exp(\Phi(\bm z_i.\bm z_p)/\tau)}}} \label{eq:sscl_i}
\end{equation}
\begin{equation}
% \mathcal{L}_{sscl} = \sum_{i = 1}^{K}{\frac{\mathcal{L}^{i}_{scl}}{\sum_{p=1}^{\vert \mathcal{P}(x_i) \vert}\alpha_{ip}}} 
\lssc = \sum_{i = 1}^{K}{\frac{\lssc^{i}}{\sum_{p}\alpha_{ip}}} 
\end{equation}

The scaling factor $\alpha_{ip}$ assigns a higher weight $s>1$ to positive samples corresponding to question paraphrases compared to other intra-class positives. Intuitively, because of the higher weight, the loss will penalize the model strongly if it fails to bring the representations of a question and its paraphrase closer. We define $\alpha_{ip}$ as: 
\begin{equation}
\label{eq:alpha}
\alpha_{ip} = 
\begin{cases}
    s & \text{if } x_p \in \xpara(x_i),   \\
    1,              & \text{otherwise}
\end{cases}
\end{equation}

\begin{algorithm}[!t]
\caption{\label{alg:alt_conclat} \concat/ with alternate \lsscl{} and \lce{}}
\begin{algorithmic}
    \STATE \textbf{input:} steps $N$; constant $N_{ce}$; data $\mathcal{D}^{\texttt{aug}}$; networks $f,g$
    % \STATE \textbf{function} CURATE($N_r$, $\mathcal{X}$, $\tau$, $\bm \alpha$)
        % \STATE $~~~~$\textcolor{gray}{\# intialize batches}        
        % \STATE $~~~~$ $\mathcal{B} = \phi, \mathcal{B}_r = \phi$
        \STATE $~~~~$ \textbf{for all} $i\in \{1, \ldots, N\}$ \textbf{do}
        \STATE $~~~~~~~~$ $\mathcal{B} = \phi$

        \STATE $~~~~~~~~$ \textbf{if} $i \Mod{N_{ce}} = 0$ \textbf{do}
        \STATE $~~~~~~~~~~~$\textcolor{gray}{\# sscl iteration}        
        \STATE $~~~~~~~~~~~$$\mathcal{B}=$ CURATE($N_r$, $\mathcal{D}^{\texttt{aug}}$, $\bm w$); $\mathcal{L}=$ \lsscl{}
        % \STATE $~~~~~~~~$ $\mathcal{L}= \mathcal{L}_{scl}$

        \STATE $~~~~~~~~$ \textbf{else} \textbf{do}
        \STATE $~~~~~~~~~~~$\textcolor{gray}{\# ce iteration}
        \STATE $~~~~~~~~~~~$$\mathcal{B} \!\sim\! \mathcal{D}^{\texttt{aug}}$; $\mathcal{L}=$ \lce{}
        \STATE $~~~~~~~~$ update $f(.),g(.)$ networks to minimize $\mathcal{L}$ over $\mathcal{B}$
        
    \STATE \textbf{return} network $f(.)$; throw away $g(.)$
\end{algorithmic}
\end{algorithm}

\subsection{Training with \lsscl{} and \lce{}} \label{ssec:alt}
We experiment with various schemes of combining supervision from \lsscl{} and \lce{} losses. Specifically, we try -- alternate (Algorithm~\ref{alg:alt_conclat}), joint, and pretrain-finetune~\cite{Khosla2020SupervisedCL} training schemes. 

Our alternate training scheme is summarized in Algorithm \ref{alg:alt_conclat}. Specifically, given $N$ total training iterations, we update our model with \lsscl{} after every $N_{ce}-1$ updates with \lce{}, where $N_{ce}$ is a hyper-parameter 
% \am{We can remove this hyperparameter, it caused some trouble in our previous reviews}
% \yk{We are not mentioning the exact value of this hyperparam}
. In the joint training scheme, we curate loss-specific batches for \lsscl{} and \lce{} but jointly update the model by accumulating the gradients of these two losses. Training alternately or jointly with the two losses simplifies the optimization procedure compared to two-stage training (pretrain-finetune as in \cite{Khosla2020SupervisedCL}) which requires double the hyper-parameters and longer training iterations. Please refer to supplementary for the exact algorithms of joint and pretrain-finetune training schemes.
% We also experiment with joint-training using a linear combination of the \lscl/ and \lce/ losses. The key drawback of this paradigm is that it does not allow for loss-specific curation of batches. Due to this, \lce/ is forced to operate with batches created for \lscl/ (built with positives and negatives). 
Empirically, we see that alternate training works slightly better than joint-training, and much better than pretrain-finetune training approach. Figure~\ref{fig:approach} depicts our training strategy (\concat/).

\subsection{Negative Types and Batch Creation} \label{ssec:batchmining}
SCL operates with multiple negative samples. 
% We define a set of negatives $\mathcal{N}(x) \subseteq \mathcal{X}^{aug}$ for any reference sample $x \in \mathcal{X}^{aug}$ as a collection of samples that have a different answer than the sample:
For a given reference sample $x = (v, q, a) \in \mathcal{D}^\texttt{aug}$, we define a corresponding set of negatives as samples with ground truth different than the reference $x$:
\begin{equation*}
    \xneg(x) = \{ (\bar{v}, \bar{q}, \bar{a})\in \mathcal{D}^\texttt{aug} \mid  \bar{a}\neq a \}
\end{equation*}

% The minibatches for \lscl{} in~\cite{Khosla2020SupervisedCL} are created by first sampling a set of anchor samples and the corresponding positive pairs. For every sample, other samples in the minibatch with a different ground-truth label is considered negative. 

We carefully curate batches for \lsscl{} by sampling different types of negatives. We classify a negative sample $\bar{x} = (\bar{v}, \bar{q}, \bar{a} ) \in \xneg(x)$ into one of three negative categories defined below.
% \ha{I think all the notation below is unnecessary. We can just simply define the three type of negatives. Because we describe how the negatives are created in text, mathematical representation seems unnecessary. Also, because we also mention how each of the negative types are denoted mathematically, equation 10 and 11 is unnecessary.}
% from $\mathcal{T}=$ \{\texttt{que, img, rand}\} via a mapping function $\Omega:{\mathcal{D} \times \mathcal{D}} \rightarrow \mathcal{T}$ such that:
% \begin{equation}
% \Omega({x}, \bar{x}) = 
% \begin{cases}
%     \texttt{img}  & \text{if } v = \bar{v}  \\
%     \texttt{que}  & \text{if } sim(q, \bar{q}) > \epsilon  \\
%     \texttt{rand} &  \text{else}
% \end{cases}
% \end{equation}

% where $\epsilon$ is a similarity threshold. 

% We partition $\xneg(x)$ in three mutually exclusive subsets $\xneg_t(x), t \in \mathcal{T}$ defined as:
% \begin{equation}
%     \xneg_t(x) = \{\bar{x} \in \xneg(x) \mid \Omega({x}, \bar{x}) = t \}
% \end{equation} 

\begin{itemize}
    \item \textbf{Image Negatives}, $ \xneg_{\texttt{img}}(x)$: Image negatives are samples 
    that have the same image ($v = \bar{v}$) as the reference ($x$) but different answer. Since VQA dataset has multiple questions ($\sim$ 5.4) per image, finding image negatives is trivial. 
    \item \textbf{Question Negatives}, $ \xneg_{\texttt{que}}(x)$:  Question negatives are samples that have questions similar to the reference but different answer. We measure the similarity between the questions by computing their cosine distance in the vector space of the Sentence-BERT \cite{reimers2019sentencebert} model, i.e. $sim(q, \bar{q}) > \epsilon$, where $\epsilon$ is a similarity threshold.
    \item \textbf{Random Negatives}, $ \xneg_{\texttt{rand}}(x)$: Random negatives are samples that do not fall under either Image or Question negative categories \ie any image and question pair that has a different answer than the reference.
\end{itemize}

\begin{algorithm}[!t]
\caption{\label{alg:batch} Batch Curation Strategy for \lsscl{}}
\begin{algorithmic}
    \STATE \textbf{input}: number of references $N_r$; data $\mathcal{D}$; weights $\bm w$
    \STATE \textbf{function} CURATE($N_r$, $\mathcal{D}$, $\bm w$)
        \STATE $~~~~$ $\mathcal{B} = \phi, \mathcal{B}_r = \phi$ \textcolor{gray}{~~~~~~~~~~~~~~~~~~~~~~~~~~~~~~~~\# initialize batches}
        \STATE $~~~~$ \textbf{for all} $i\in \{1, \ldots, N_r\}$ \textbf{do}
        \STATE $~~~~~~~~$$x_i  \!\sim\! \mathcal{D}$ \textcolor{gray}{~~~~~~~~~~~~~~~~~~~~~~~~~~~~~~~~~~~~~~~~~~~~~~~~~~~~~~\# reference}
        %\STATE $~~~~~~~~$\textcolor{gray}{\# sample intra-class positive}        
        \STATE $~~~~~~~~$$\hat{x_i}  \!\sim\! \xcls(x_i)$ \textcolor{gray}{~~~~~~~~~~~~~~~~~~~~~~~~~~~~~\# intra-class positive}
        % \STATE $~~~~~~~~$\textcolor{gray}{\# pick negative type}        
        \STATE $~~~~~~~~$$t  \!\sim\! \text{Cat}(\mathcal{T} \vert \bm w)$ \textcolor{gray}{~~~~~~~~~~~~~~~~~~~~~~~~~~~~~~~~~~~~~\# negative type}
        % \STATE $~~~~~~~~$\textcolor{gray}{\# sample negative from correspoding set}        
        \STATE $~~~~~~~~$$\bar{x}_i  \!\sim\!  \xneg_{t}(x_i)$ \textcolor{gray}{~~~~~~~~~~~~~~~~~~~~~~~~~~~~~~~~~~~~~~~~~~~~~~~~\# negative}
        \STATE $~~~~~~~~$append $\mathcal{B} = \mathcal{B} \cup \{x_i, \hat{x_i}, \bar{x}_i\}$\\

        \STATE $~~~~$ \textbf{for all} $i\in \{1, \ldots, \vert \mathcal{B} \vert\}$ \textbf{do}
        %\STATE $~~~~~~~~$\textcolor{gray}{\# sample paraphrased positive}        
        \STATE $~~~~~~~~$${x'_i}  \!\sim\! 
        \xpara(x_i)$ \textcolor{gray}{~~~~~~~~~~~~~~~~~~~~~~~\# paraphrased positive}
        \STATE $~~~~~~~~$append $\mathcal{B}_r = \mathcal{B}_r \cup \{x'_i\}$
    \STATE \textbf{return} $\mathcal{B} \cup \mathcal{B}_r$
\end{algorithmic}
\end{algorithm}

We hypothesize that discriminating between joint V+L representations of above negatives and the reference would lead to more robust V+L representations as it requires the model to preserve relevant information from both modalities in the learnt representation. Negative samples belonging to each of the above types are depicted in Figure \ref{fig:approach}(b). \smallskip

\noindent \textbf{Batch Curation}. To create mini-batches for \lsscl/, as described in Algorithm~\ref{alg:batch}, we start by filling our batch with triplets of reference $x_i$, a intra-class positive $\hat{x_i}$ and a negative sample $\bar{x_i}$ of type $t$. The negative type $t$ is sampled from a categorical distribution $\text{Cat}(\mathcal{T} \vert\bm w)$ where $\bm w = (w_{\texttt{img}}, w_{\texttt{que}}, w_{\texttt{rand}})$ are the probability weights of selecting different types of negatives defined by $\mathcal{T}=$ \{\texttt{que, img, rand}\}. This procedure is repeated for specified number of times $N_r$ to create a batch $\mathcal{B}$. Finally, for every sample in $\mathcal{B}$ we add a corresponding paraphrased positive ${x'_i}$ sample. 
% When sampling all the three types of negatives we use $\bm w = (w_{\texttt{img}}, w_{\texttt{que}}, w_{\texttt{rand}}) = (0.25, 0.25, 0.5)$. 
For \lce{}, we sample mini-batches randomly from the dataset $\mathcal{D}^\texttt{aug}$. \smallskip 

\noindent \textbf{Importance of Scaling Factor}. VQA Dataset has a skewed distribution of answer labels and since we sample references for \SCL/ minibatch independently of each other (see Algorithm~\ref{alg:batch}) quite often we end up with multiple intra-class positives but only a single paraphrased positive for given a reference in a minibatch. To balance this trade-off we choose to scale the loss corresponding to paraphrased positive sample from the intra-class positive samples. We call this loss Scaled Supervised Contrastive Loss (\lsscl/).

%% file: sections/experiments.tex
\section{Experiments}
\input{tables/iccv_ablations}

\vspace{5pt}
\subsection{Datasets and Metrics} We use the VQA v2.0 \cite{goyal2016making} and the VQA-Rephrasings~\cite{shah2019cycle} datasets for experiments. VQA contains nearly 443K train, 214K val and 453K test instances. VQA-Rephrasings was collected to evaluate the robustness of VQA models towards human rephrased questions. Specifically, the authors collected 3 human-provided rephrasings for 40k image-question pairs from the VQA v2.0 validation dataset. 

Shah \etal\cite{shah2019cycle} also introduced Consensus Score (CS) as an evaluation metric to quantify the agreement of VQA models across multiple rephrasings of the same question. Amongst all subsets of paraphrased questions of size k, the consensus score \textbf{CS(k)} measures the fraction of subsets in which \emph{all} the answers have non-zero VQA-Score. For a set of paraphrases $Q$, the consensus score \textbf{CS(k)} is defined as:

\begin{equation}
\label{eq:consesus_score}
\textbf{CS(k)} \quad = \sum_{Q' \subset Q, |Q'|=k} \frac{\mathcal{S}(Q^\prime)}{\Mycomb{k}}
\end{equation}

\begin{equation}
\label{eq:all_correct}
\mathcal{S}(Q^\prime) = \begin{cases}
        \quad  1 & \text{if } \enskip \forall q \in Q^\prime, \enskip \text{VQA-Score}(q) > 0, \\
        \quad  0 & \text{else}
\end{cases}
\end{equation}

Where $\Mycomb{k}$ is number of subsets of size $k$ sampled from a set of size $n$.
$\textbf{CS(k)}$ is zero for a group of questions $Q$ when the model answers at least $k$ questions correctly.

When reporting results on the val split and VQA-Rephrasings, we train on the VQA 2.0 train split and when reporting results on the VQA 2.0 test-dev and test-std we train on both VQA 2.0 train and val splits. The VQA Rephrasings dataset \cite{shah2019cycle} is never used for training and used only for evaluation. 

\subsection{Baselines and Training Details}
\textbf{VQA Model}. For $f$, we use a multimodal transformer (MMT) inspired from %\uni/
~\cite{chen2019uniter}, with 6 layers and 768-dim embeddings. It takes as input two different modalities. The question tokens are encoded using a pre-trained three layer BERT~\cite{devlin2018bert} encoder which is fine-tuned along with the multimodal transformer. Object regions are encoded by extracting features from a frozen ResNeXT-152~\cite{Xie_2017} based Faster R-CNN model~\cite{ren2015faster}. The projection module $g$ consists of two linear layers and a L-2 normalization function. We choose MMT as representative of current SoTA models~\cite{Jiang2020InDO, lu2019vilbert, chen2019uniter, li2020oscar, mcb} in VQA that rely heavily on some form of multi-modal transformer architecture. Also note that our approach (\concat/) is \textit{agnostic} to the choice of the model.

\noindent \textbf{Question Paraphrases using VQG}. Apart from training with question paraphrases generated via \BT/, we also experiment with generating question paraphrases using the VQG module from \cite{shah2019cycle}. We input the VQG module with 88 random  noise vectors to keep the generation comparable with \BT/ approach. For filtering, we use the gating mechanism used by the authors and sentence similarity score of $\geq 0.85$ and keep a maximum of 3 unique rephrasings for each question.\smallskip

\noindent \textbf{Training Details}. We train our models using Adam optimizer \cite{adam} with a linear warmup and with a learning rate of 1e-4 
and a staircase learning rate schedule, where we multiply the learning rate by 0.2 at 10.6K and at 15K iterations. We train for 5 epochs of train + augmented dataset on 4 NVIDIA Titan XP GPUs and use a batch-size of 420 when using \lsscl/ and \lce/ both and 210 otherwise. We put remaining hyperparameters in the supplementary. \smallskip

\noindent \textbf{Existing state-of-the-art methods}. Previous work \cite{shah2019cycle} in VQA-Rephrasings trained a VQG model using a cycle-consistent training scheme along with the VQA model. The approach involved generating questions by a VQG model such that the answer for the original and the generated question are consistent with each other. For their experiments, they build on top of Pythia~\cite{jiang2018pythia} and BAN~\cite{bua} as base VQA models. We treat these approaches as baselines for our experiments.

\input{tables/iccv_main}

%% file: tables/iccv_ablations.tex
\newcolumntype{H}{>{\setbox0=\hbox\bgroup}c<{\egroup}@{}}
\begin{table*}[h!]\centering
\setlength{\tabcolsep}{3.5pt}
\begin{tabular}{clcccHccccHH}
 \toprule
  \multirow{2}{*}{} & \multirow{2}{*}{\textbf{Model}} & \multirow{2}{*}{\textbf{Loss(es)}} & \multirow{2}{*}{\textbf{Scaling}}& \multirow{2}{*}{\textbf{N-Type}} & \multicolumn{1}{c}{\textbf{}} & \multirow{2}{*}{\textbf{Train Scheme}} & \multirow{2}{*}{\textbf{CS(3)}} & \multirow{2}{*}{\textbf{CS(4)}} & \multicolumn{3}{c}{\textbf{VQA}} \\

  ~ &  ~ &  ~ & ~ & ~ & \textbf{k=1} & \textbf{~} & ~ & ~ & \textbf{val} & \textbf{test-dev}&  \textbf{test-std} \\
 \midrule
 \band \small\texttt{1} & MMT   & \lce{} & - & - & - & - & 55.53 & 52.36 & 66.31 & 69.51 & 69.22 \\
  \small\texttt{2} & MMT   & \lsscl{} \& \lce{} & \ding{51} & \texttt{R} & 68.19 & Alternate & 56.53 & 53.42 & 66.62 & - & -\\
\band \small\texttt{3} & MMT  & \lscl{} \& \lce{} & \ding{51} & \texttt{RQ} & 68.41 & Alternate & 56.88 & 53.77 & 66.97 & - & -\\
 \small\texttt{4} &  MMT   & \lsscl{} \& \lce{} & \ding{51} & \texttt{RI} & 68.47 & Alternate & 56.91 & 53.79 & 66.93 & - & -\\
\band \small\texttt{5} & MMT   & \lsscl{} \& \lce{} & \ding{51} & \texttt{QI} & {68.65} & {Alternate} & {57.00} & {53.90} & {66.95} & - & - \\
\small\texttt{6} & MMT   & \lsscl{} \& \lce{} & \ding{51} & \texttt{RQI} & \textbf{68.62} & Alternate & \textbf{57.08} & \textbf{53.99} & \textbf{66.98} & {69.80} & \textbf{70.00} \\

\midrule
\band \small\texttt{7} & MMT   & \lscl{} \& \lce{} & \xmark & \texttt{RQI} & 68.20 & Alternate & 56.49 & 53.36 & 66.60 & - & -\\

\small\texttt{8} & MMT & \lsscl{} \& \lce{} & Dynamic (Eq.~\ref{eq:alpha-dynamic}) & \texttt{RQI} & 68.60 & Alternate & 57.01 & 53.92 & 66.95 & - & - \\

\midrule
\band \small\texttt{9} & MMT  & \lsscl{} \& \lce{} & \ding{51} & \texttt{RQI} & 66.95 & Joint & 56.59 & 53.63 & 66.23 & - & - \\

 \small\texttt{10} & MMT   & \lsscl{} $\rightarrow$ \lce{}~\cite{Khosla2020SupervisedCL} & \xmark & \texttt{RQI}& ~ & Pretrain-Finetune & 52.63 & 49.20 & 64.21 & - & - \\

\midrule

 \band \small\texttt{11} & MMT  & $\mathcal{L}_{\text{DMT}}$~\cite{free_vqa} \& \lce{} & \xmark & \texttt{RQI}& ~ & Alternate & 56.23 & 53.10 & 66.59 & 68.28 & 68.38 \\  

\bottomrule

\end{tabular}
  \caption{\textbf{Ablations Study}. \textbf{Scaling} denotes whether scaling factor $\alpha$ (defined in Eq.~\ref{eq:alpha} or Eq.~\ref{eq:alpha-dynamic}) was used. \textbf{N-Type} defines the type of negatives used from Image (\texttt{I}), Question (\texttt{Q}) and Random (\texttt{R}). All experiments are run with \BT/ data.}
  \vspace{-10pt}
  \label{ablation_table}
\end{table*}

%% file: tables/iccv_main.tex
\begin{table}[t] \footnotesize
\setlength\tabcolsep{3 pt}
\resizebox{\columnwidth}{!}{
  \begin{tabular}{l l c c c c  c c}
  \toprule
  \multirow{2}{*}{} &\multirow{3}{*}{Model} & \multirow{3}{*}{DA} & \multicolumn{2}{c}{Consensus Scores} & \multicolumn{3}{c}{VQA Scores} \\  
  \cmidrule(lr){4-5}
  \cmidrule(lr){6-8}
  & & & CS(3) & CS(4) & val & test-dev & test-std\\
  \midrule
   %\multirow{2}{*}{\small{VQA}} 
\band \footnotesize\texttt{1} & Pythia~\cite{jiang2018pythia} & - & 45.94	& 39.49	&65.78	& 68.43 & -  \\
\footnotesize\texttt{2} & BAN~\cite{kim2018bilinear} & - & 47.45	& 39.87	& 66.04	& 69.64 & -  \\
\band \footnotesize\texttt{3} & Pythia + CC~\cite{shah2019cycle}  & - & {50.92}	& {44.30}	& {66.03}	& 68.88 & -\\
\footnotesize\texttt{4} & BAN + CC~\cite{shah2019cycle} & - & {51.76}	& {48.18}	& {66.77}	& \textbf{69.87} & -\\
  \midrule
\band \footnotesize\texttt{5} & MMT & - & 55.10 & 51.82 & 66.46 & - & - \\
  
\footnotesize\texttt{6} & MMT & VQG~\cite{shah2019cycle} & 54.92 & 51.85 & 64.50 &  - & -\\
\band \footnotesize\texttt{7} & MMT + \concat/ & VQG~\cite{shah2019cycle} & 55.33 & 52.31 & 64.74 & - & - \\
\footnotesize\texttt{8} & MMT & BT & 55.53 & 52.36 & 66.31 & 69.51 & 69.22  \\
\band \footnotesize\texttt{9} & MMT + \concat/ & BT & \textbf{57.08} & \textbf{53.99} & \textbf{66.98} & {69.80} & \textbf{70.00}  \\
  \bottomrule
  \end{tabular}}
    \smallskip
    \caption{\concat/ vs existing methods / baselines on VQA-Rephrasings and VQA 2.0. \textbf{DA} denotes the source of augmented data from either Back Translation (BT) or Visual Question Generation (VQG). For test-dev and test-std, we train our model on train+val set of VQA 2.0.}
  \vspace{10pt}
 \label{main_table}
\end{table}

%% file: sections/results.tex
\section{Results}
In this section, we carefully ablate each component of \concat/, and also compare results with previous methods (Pythia+CC, BAN+CC) from~\cite{shah2019cycle}. We report the Consensus Score (\textbf{CS(k)}) for $k={3,4}$ on VQA-Rephrasings~\cite{shah2019cycle} and VQA Accuracy on VQA 2.0~\cite{goyal2016making} datasets. We omit \textbf{CS(1)} and \textbf{CS(2)} for brevity, and provide them in the supplementary.  

% \am{Why not 1 and 2 as well in the ablations table for completeness?}.

\subsection{\concat/}
Our baseline architecture MMT without any additional data (Table~\ref{main_table}, Row 5) and trained using cross-entropy (\lce{}) outperforms previous best (BAN+CC, Table~\ref{main_table}, Row 4) by +3.64\% on \textbf{CS(4)} while being -0.31\% worse on VQA 2.0 validation. Training MMT with Back-translated data (Table~\ref{main_table}, Row 8) using only \lce{} further improves \textbf{CS(4)} by +0.54\% while slightly degrading performance on VQA 2.0 by -0.15\%, we treat this as our new baseline.     

We find that alternate training (\concat/) with \lsscl/ and \lce/ (Table~\ref{main_table}, Row 9) improves \textbf{CS(4)} by +1.63\% and VQA Accuracy by +0.67 \% on validation. \concat/ outperforms previous state-of-the-art approach BAN+CC by +5.81\% on \textbf{CS(4)} while performing competitively on VQA 2.0 validation (+0.22\%) and test-dev (-0.07\%) splits. We present this as our main result, which shows that training with both the losses together leads to models that are accurate (higher VQA score) and robust (higher Consensus score). 

\noindent\textbf{\concat/ with VQG data}. We also experiment by augmenting the data generated from VQG model of~\cite{shah2019cycle}. Similar to \BT/ data, we find that using \concat/ (Table~\ref{main_table}, Row 7) leads to +0.46\% and +0.24\% gains on \textbf{CS(4)} and VQA 2.0 validation over the baseline (Table~\ref{main_table}, Row 8). We attribute the relatively smaller gains from VQG data to the lower quality and lesser quantity of paraphrases generated by the VQG module. We discuss more about the quality of generated data in Supplementary Section 5.

% Whitehead \etal\cite{whitehead2020learning} report +0.4\% gains on VQA-Rephrasings~\cite{shah2019cycle} on \textbf{CS(4)} using a modular architecture, whereas, we show much stronger gains (+1.63 \%) while also being agnostic to the model architecture.

\subsection{Ablations}
\noindent\textbf{Training schemes}. We try three different ways of combining \lce{} and \lsscl{} losses. Training alternately performs the best (Table~\ref{ablation_table}, Row 6), whereas training jointly performs worse by -0.36\% and -0.75\% on \textbf{CS(4)} and VQA validation accuracy respectively (Table~\ref{ablation_table}, Row 9). Following the approach taken in~\cite{Khosla2020SupervisedCL}, we try pre-training the model with \lsscl/ and then finetuning it on \lce/ (Table~\ref{ablation_table}, Row 10) and we find this to perform the worst with -4.79\% and -2.77\% in \textbf{CS(4)} and VQA validation accuracy respectively
% \am{Needs polishing in language. Devi: Seems fine to me...}
.

\begin{figure*}[t!]
  \centering
  \includegraphics[width=0.97\textwidth]{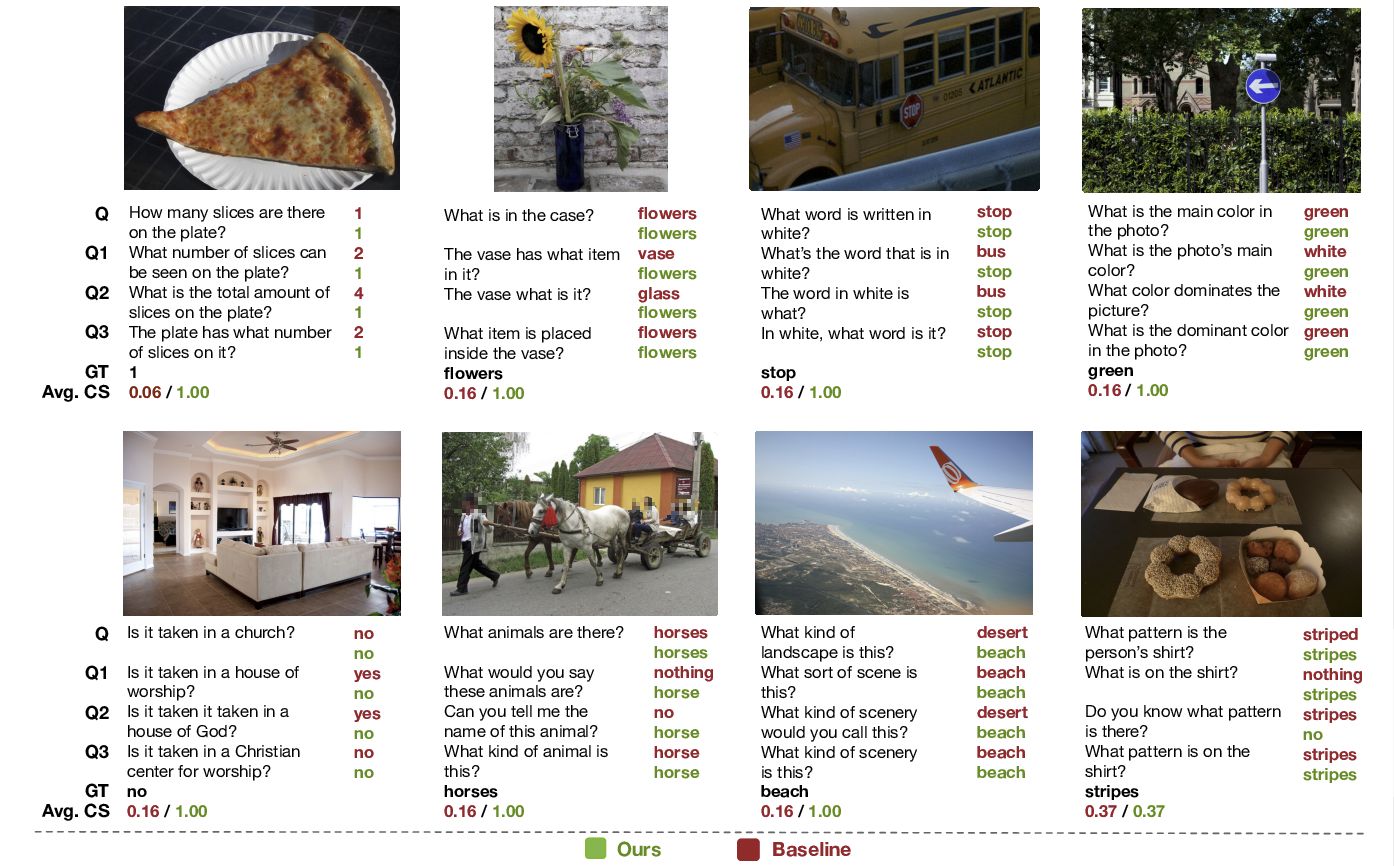}
  \caption{\textbf{Qualitative Examples.} Predictions of \concat/ (Table~\ref{ablation_table}, Row 1) and our baseline (Table~\ref{ablation_table}, Row 6) on several image-question pairs and their corresponding rephrased questions. Average Consensus Scores (k=1-4) are also shown at the bottom (higher the better).}
  \label{fig:qualitative}
  \vspace{-15pt}

\end{figure*}

\noindent\textbf{Contrastive vs Triplet Losses}. %VQA has a skewed distribution of answer labels and since we sample reference samples for \lscl/ minibatch independently of each other (see Algorithm~\ref{alg:batch}) quite often we end up with multiple intra-class positives but only a single paraphrased positive for given a reference sample in a minibatch. To balance this trade-off we choose to scale the loss corresponding to paraphrased positive sample, we call this loss Scaled Supervised Contrastive Loss (\lsscl/).
% Compared to Supervised Contrastive Loss \lscl/~\cite{Khosla2020SupervisedCL} (Row 15), we also see improvement on both VQA Accuracy and Consensus Scores when using our proposed variant Scaled Supervised Contrastive Loss \lsscl/ (Row 16). We find that using \lsscl/ is more effective when used with contrastive batches (Row 16 vs Row 12).
Previous works have explored the use of triplet losses~\cite{free_vqa, patro2018differential} for learning robust VQA models. Specifically, we experiment by replacing our \lsscl/ with Dynamic-margin Triplet loss ($\mathcal{L}_{\text{DMT}}$) proposed in~\cite{free_vqa} for mitigating the tendency of VQA models to ignore the image and rely solely on question for answering (also known as knowledge-inertia)
% \am{knowledge-intertia bracket can be removed}
. It is also worth noting that $\mathcal{L}_{\text{DMT}}$ is an improved version of the vanilla triplet loss used in~\cite{patro2018differential}. We find that \concat/ outperforms this ablation (Table~\ref{ablation_table}, Row 11) by +0.89\% and +0.39\% \textbf{CS(4)} and VQA validation accuracy respectively.  

% \textbf{Negative Sampling for \lsscl/}. We categorize the negatives into three non-overlapping categories based on the similarity w.r.t. reference question and image. By adding samples in which images are different (\texttt{img}-type) and samples that have different answers for very similar quesitons (\texttt{ques}-type) as negatives, it helps learn more robust V+L representations since it requires the model to preserve relevant information of both modalities in the learned representation. We also confirm this via empirical results, specifically, sampling often from \texttt{img}-type (Row 13),  \texttt{que}-type (Row 14) and both types (Row 16) provides strong gains on both VQA accuracy and Consensus Scores.       

% \textbf{Back-Translation(BT) vs Visual Question Generation (VQG)}. We also ablate on choice of our data augmentation strategy, and show the efficacy of our approach using on both back-translation (BT) and \vqg/ (VQG)\cite{shah2019cycle}. We find the quality and diversity of paraphrases generated by the VQG module to be poor. Nonetheless, we show that using \concat/ with VQG rephrasings leads to gains on both VQA and Consensus Scores (Row 6,7).  

\noindent\textbf{Scaling in \lsscl{}}. We see improvement on both VQA validation (+0.56\%) and \textbf{CS(4)} (+0.35\%) when using our proposed variant Scaled Supervised Contrastive Loss (\lsscl/) when compared to using unscaled \lscl/ (Table~\ref{ablation_table}, Rows 6, 7). Beyond the constant scaling factor defined in Eq.~\ref{eq:alpha}, we also experimented with using a dynamic scaling factor defined as follows:
    \begin{equation}
    \label{eq:alpha-dynamic}
    \alpha_{ip} = 
    \begin{cases}
        s.\Phi(\bm z_i.\bm z_p) & \text{if } x_p \in \xpara(x_i),   \\
        \Phi(\bm z_i.\bm z_p),              & \text{otherwise}
    \end{cases}
    \end{equation}

Where $\mathrm{\Phi}(\bm u,\bm v) = 1 - \bm u^\top \bm v / \lVert\bm u\rVert \lVert\bm v\rVert$ computes the cosine distance between $\bm u$ and $\bm v$. We did not find significant improvements using dynamic scaling (Table~\ref{ablation_table}, Row 8).

%We hypothesize that adding these negative samples encourage the model to learn robust V+L representations by preserving relevant information from both modalities in the learned representation.\smallskip

\noindent\textbf{Negative Sampling Strategy}. Furthermore, we find that our proposed negative sampling strategy (Algorithm~\ref{alg:batch}) where we carefully curate batches for \lsscl/ loss (Table~\ref{ablation_table}, Row 6) helps improve \textbf{CS(4)} (+0.57\%) and VQA accuracy (+0.36\%) over random-sampling (Table~\ref{ablation_table}, Row 2). We find that adding either \texttt{que}-type negatives (Table~\ref{ablation_table}, Row 3) or \texttt{img}-type negatives(Table~\ref{ablation_table}, Row 4) lead to gains in \textbf{CS(4)} and VQA validation accuracy. Using only \texttt{img}-type and \texttt{que}-type negatives (Table~\ref{ablation_table}, Row 5) leads to significant gains, showing that use of both the types is crucial.
% \ha{Say something about Row 5 as well}
%We hypothesize that adding these negative samples encourage the model to learn robust V+L representations by preserving relevant information from both modalities in the learned representation.\smallskip

\subsection{Qualitative Analysis}
We qualitatively visualize few samples in Figure~\ref{fig:qualitative}. We compare our final approach (Table~\ref{ablation_table}, Row 6) with our baseline (Table~\ref{ablation_table}, Row 1). As evident from samples, \concat/ improves the consistency in answers across the rephrasings. (2,2) shows an interesting example where \concat/ yields a singular answer for one question paraphrase and produces the original plural answer for other paraphrased question.
In (2,3), baseline incorrectly answers the original VQA question but correctly answers some of the rephrasings whereas our approach gets all the questions right. (2,4) illustrates a failure case where both the approaches fail to answer all the paraphrased questions correctly.

\section{Conclusion}
To summarize, we have three main contributions. First, we propose a novel training paradigm (\concat/) that optimizes contrastive and \ce/ losses to learn joint vision and language representations that are robust to question paraphrases. Minimizing the contrastive loss encourages representations to be robust to linguistic variations in questions while the \ce/ loss preserves the discriminative power of the representations for answer classification. Second, we introduce Scaled Supervised Contrastive Loss (\lsscl{}), that assigns higher weight to positive samples associated with question paraphrases over samples that just have the same answer boosting the performance further. Finally, we propose a negative sampling strategy to curate loss-specific batches which improves performance over random sampling strategy. Compared to previous approaches, VQA models trained with \concat/ achieve higher consistency scores on the VQA-Rephrasings dataset as well as higher VQA accuracy on the VQA 2.0 dataset across a variety of data augmentation strategies. We also qualitatively demonstrate that our approach yields correct and consistent answers for VQA questions and their rephrasings.

\section{Acknowledgements}
We thank Abhishek Das, Prithvijit Chattopadhyay and Arjun Majumdar for their feedback. The Georgia Tech effort was supported in part by NSF, AFRL, DARPA, ONR YIPs, ARO PECASE, Amazon. The views and conclusions contained herein are those of the authors and should not be interpreted as necessarily representing the official policies or endorsements, either expressed or implied, of the U.S. Government, or any sponsor.

%% file: supp-sections/supp_experiments.tex
\section{Ablations with Joint Training}
In the joint training experiment (Table 2, Row 8), we use a weighing parameter ($\beta$) to combine the \lscl/ and \lce/ losses. We ablate on the choice of weight ($\beta$) used, and we represent the overall loss in this experiment as: 
$$    \mathcal{L}_{joint} = \beta\lssc + (1-\beta)\mathcal{L}_{\text{CE}} $$

We also find that the VQA-Accuracy and Consensus Scores hit a sweet-spot at $\beta=0.5$ and we use this configuration as our basline. 
\input{tables/joint_ablate}

\section{Joint and Pretrain-Finetune Training}
As mentioned in Section 4.3 of the manuscript, we respectively provide the training schemes used to jointly optimize in Algorithm~\ref{alg:joint_conclat} and the scheme used to pretrain-finetune in Algorithm~\ref{alg:pre_conclat} with the \lsscl{} and \lce{} losses.

\begin{algorithm}[t]
\caption{\label{alg:joint_conclat} \concat/ with joint \lsscl{} and \lce{}}
\begin{algorithmic}
    \STATE \textbf{input:} steps $N$; constant $N_{r}, \beta$; data $\mathcal{D}^{\texttt{aug}}$; networks $f,g$
    % \STATE \textbf{function} CURATE($N_r$, $\mathcal{X}$, $\tau$, $\bm \alpha$)
        % \STATE $~~~~$ $\mathcal{B} = \phi, \mathcal{B}_r = \phi$
        \STATE $~~~~$ \textbf{for all} $i\in \{1, \ldots, N\}$ \textbf{do}
        % \STATE $~~~~~~~~$ $\mathcal{B} = \phi$
        % \STATE $~~~~~~~~$ \textbf{if} $i \Mod{N_{ce}} = 0$ \textbf{do}
        \STATE $~~~~~~~~$\textcolor{gray}{\# initialize batches}        
        \STATE $~~~~~~~~$$\mathcal{B}_{\text{SSC}}=$ CURATE($N_r$, $\mathcal{D}^{\texttt{aug}}$, $\bm w$); $\mathcal{B}_{\text{CE}} \!\sim\! \mathcal{D}^{\texttt{aug}}$
        \STATE $~~~~~~~~$\textcolor{gray}{\# compute gradients separately}   
        \STATE $~~~~~~~~$$\nabla_{\text{SSC}}= \nabla$ \lsscl{}$(f, g, \mathcal{B}_{\text{SSC}}) \cdot \beta$
        \STATE $~~~~~~~~$$\nabla_{\text{CE}}= \nabla$ \lce{}$(f, g, \mathcal{B}_{\text{CE}}) \cdot (1-\beta)$
        \STATE $~~~~~~~~$\textcolor{gray}{\# joint update}   
        \STATE $~~~~~~~~$ update $f(.),g(.)$ networks with $\nabla = \nabla_{\text{CE}} + \nabla_{\text{SSC}}$
        
    \STATE \textbf{return} network $f(.)$; throw away $g(.)$
\end{algorithmic}
\end{algorithm}

\begin{algorithm}[t]
\caption{\label{alg:pre_conclat} \concat/ with pre-train \lsscl{} and fine-tune \lce{}}
\begin{algorithmic}
    \STATE \textbf{input:} steps $N_p, N_f$; data $\mathcal{D}^{\texttt{aug}}$; networks $f,g$
        \STATE $~~~~$\textcolor{gray}{\# pretrain with SSCL}        
        \STATE $~~~~$ \textbf{for all} $i\in \{1, \ldots, N_p\}$ \textbf{do}
        \STATE $~~~~~~~~$$\mathcal{B}=$ CURATE($N_r$, $\mathcal{D}^{\texttt{aug}}$, $\bm w$)
        \STATE $~~~~~~~~$ update $f(.),g(.)$ networks to minimize \lsscl{} over $\mathcal{B}$
        
        \STATE $~~~~$\textcolor{gray}{\# finetune with CE}        
        \STATE $~~~~$ \textbf{for all} $i\in \{1, \ldots, N_f\}$ \textbf{do}
        \STATE $~~~~~~~~$$\mathcal{B} \!\sim\! \mathcal{D}^{\texttt{aug}}$
        \STATE $~~~~~~~~$ update $f(.)$ network to minimize \lce{} over $\mathcal{B}$
        
    \STATE \textbf{return} network $f(.)$; throw away $g(.)$
\end{algorithmic}
\end{algorithm}

\section{Gradient Surgery of  \lsscl/ and \lce/}
\begin{figure}[t!]
  \centering
  \renewcommand{\thefigure}{A}
  \includegraphics[width=\linewidth]{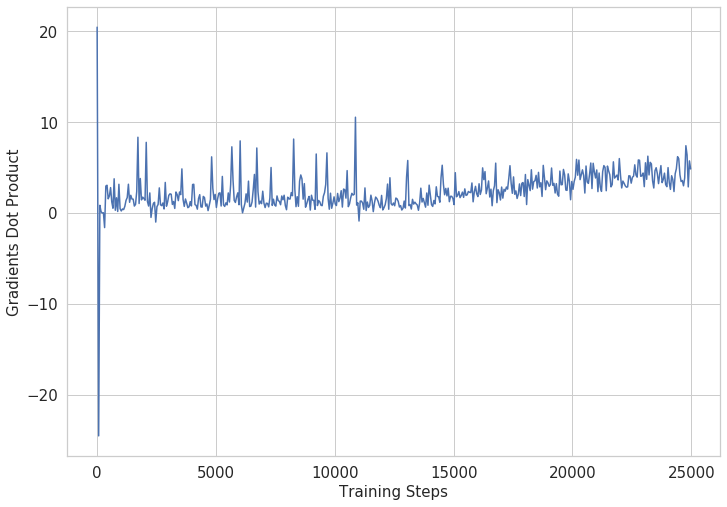}
  \caption{Gradient Alignment between the \lsscl/ and \lce/ losses. The dot-product is positive indicating that the gradients from the two losses are aligned.}
  \label{fig:pcgrad}
\end{figure}

To know whether the gradients of both the losses (\lsscl/ and \lce/) are aligned with each other during training, we follow the gradient surgery setup of~\cite{yu2020gradient} for multi-task learning. During joint-training, we take the dot-products of gradients from both the losses and plot them to see how well they are aligned \ie whether the dot product is positive or negative. In Figure \ref{fig:pcgrad} we plot the un-normalized dot product between the gradients corresponding to \lce/ and \lsscl/ losses. We find that except for initial few steps the gradients of both the losses are aligned (dot product is positive) and thus the updates are complementary with respect to each other.

\input{tables/supp_hyper}

\section{Training} 

\textbf{Hyperparameters.} All the models have $\sim$100M trainable parameters.  We train our models using Adam optimizer \cite{adam} with a linear warmup and with a learning rate of 1e-4 and a staircase learning rate schedule, where we multiply the learning rate by 0.2 at 10.6K and at 15K iterations. We train for 5 epochs of augmented train dataset $\mathcal{D}^{aug}$ on 4 NVIDIA Titan XP GPUs and use a batch-size of 420 when using \lsscl/ and \lce/ both and 210 otherwise. We use PyTorch~\cite{Paszke2019PyTorchAI} for all the experiments. Hyperparameters are summarized in Table \ref{tab:hyperparameters_supp}. 

\section{Frequently Asked Questions}

\noindent \textbf{Why use different sampling rates for different negative types?}

\noindent The different types of negatives -- same-image-different-question (\texttt{img}) and same-question-different-image (\texttt{que}) -- encourages the model to be sensitive to both modalities. We use different sampling weights to emphasize more on these two types of negatives over the ones which just have different answers. We obtain the weights (Table~\ref{tab:hyperparameters_supp}, Row 9) through hyper-parameter tuning on the validation set. \\

\noindent \textbf{Why should questions dealing with different concepts but same answer (e.g., questions in Fig 2b, ``Is the dog atop a sofa?" and ``Is there broccoli in the picture?") have similar representations?} 

\noindent We clarify that we do not impose any supervision at the level of MMT layers but only at the penultimate layer before answer prediction. Hence, the model is able to perform different reasoning steps (needed to process entirely different visual/textual inputs) for arriving at the same final answer.

\section{Augmented Data}
\textbf{Back-translation}: We use 88 different MarianNMT~\cite{mariannmt} Back-translation model pairs released by Hugging Face~\cite{Wolf2019HuggingFacesTS} to generate question paraphrases.  We  use  Sentence-BERT~\cite{reimers2019sentencebert} to filter out paraphrases that cosine similarity of $\geq 0.95$  with  the  original  question  and  choose  three  unique paraphrases randomly from the filtered set. After filtering duplicates we end up with 2.89 paraphrases per original question on average. 

\textbf{VQG}: We use the VQG model introduced by previous work \cite{shah2019cycle} that takes as input the image and answer to generate a paraphrased question. We input the VQG module with 88 random noise vectors to keep the generation comparable with \BT/ approach. For filtering, we use the gating mechanism used by the authors and sentence similarity score of $\geq 0.85$ and keep a maximum of 3 unique rephrasings for each question. Since, VQG produces fewer unique rephrasings per question than Back-translation, we used a lower similarity threshold. After filtering duplicates we end up with only 0.96 paraphrases per original question on average, far fewer than Back-translation. Qualitatively, we find the VQG paraphrases worse when compared against Back-translated ones.

\textbf{Evaluation:} During training, we evaluate our models using the Back-translated rephrasings on a subset of questions from validation set which do not overlap with VQA-Rephrasings~\cite{shah2019cycle} dataset.

\section{Code and Result Files} 
We share the code for running the baseline and the best experiments (Table 1, Rows 5, 9). Please find the released code at: \textcolor{blue}{\href{https://www.github.com/yashkant/concat-vqa}{https://www.github.com/yashkant/concat-vqa}}

\section{Full Ablations}
For brevity and conciseness, we omitted \textbf{CS(1)} and \textbf{CS(2)} scores in the main ablation table, we provide the these scores in Table~\ref{ablation_table_supp}.

\input{tables/iccv_ablations_full}

\section{Qualitative Samples}
\begin{figure*}[t!]
  \renewcommand{\thefigure}{B}
  \centering
  \includegraphics[width=0.9\textwidth]{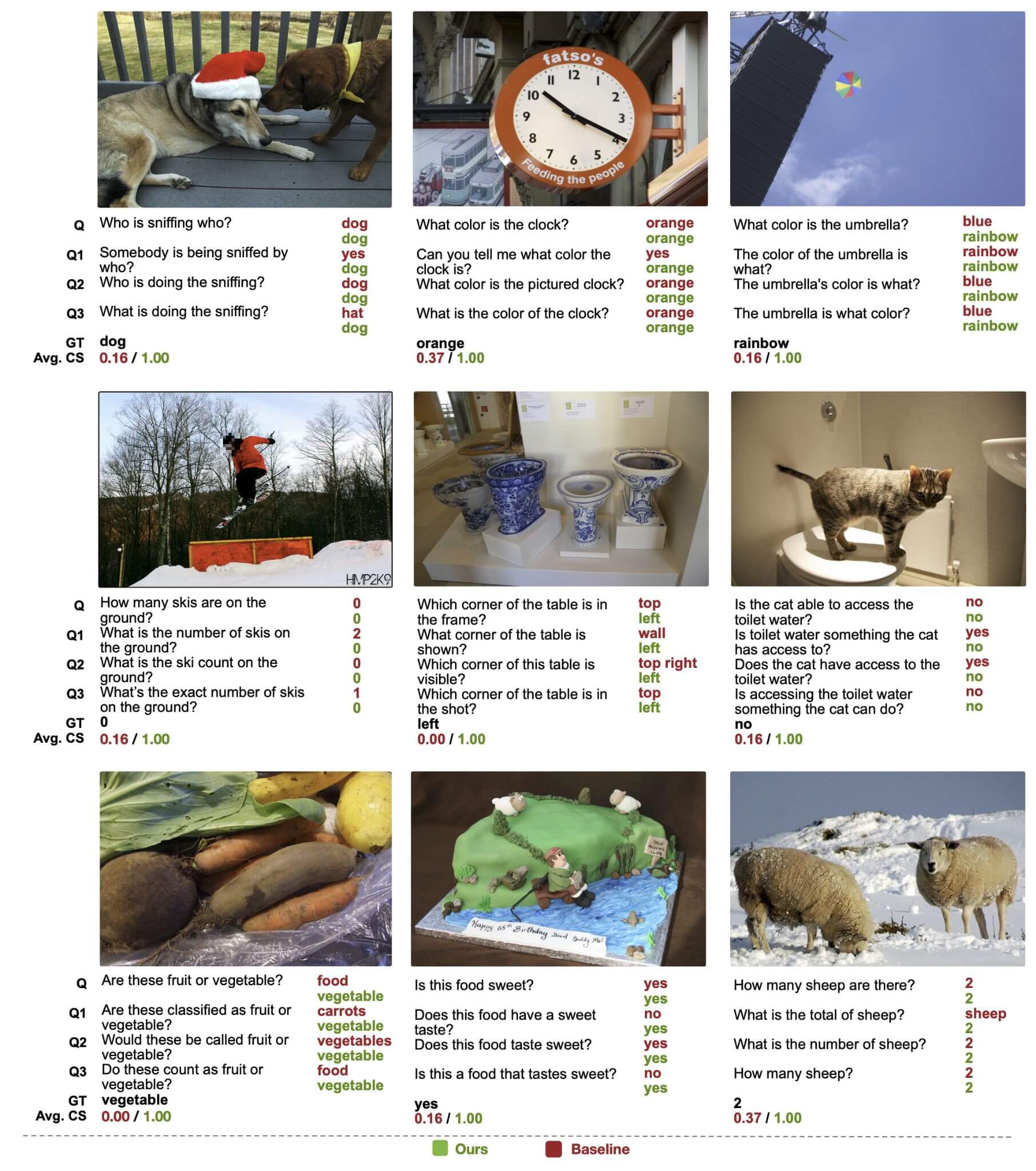}
  \caption{Qualitative Examples. Predictions of \concat/ and MMT+CE baseline on several image-question pairs and their corresponding rephrased questions. Average Consensus Scores (k=1-4) are also shown at the bottom (higher the better).}
  \label{fig:qualitative1}
\end{figure*}

\begin{figure*}[t!]
  \centering
    \renewcommand{\thefigure}{C}
  \includegraphics[width=0.9\textwidth]{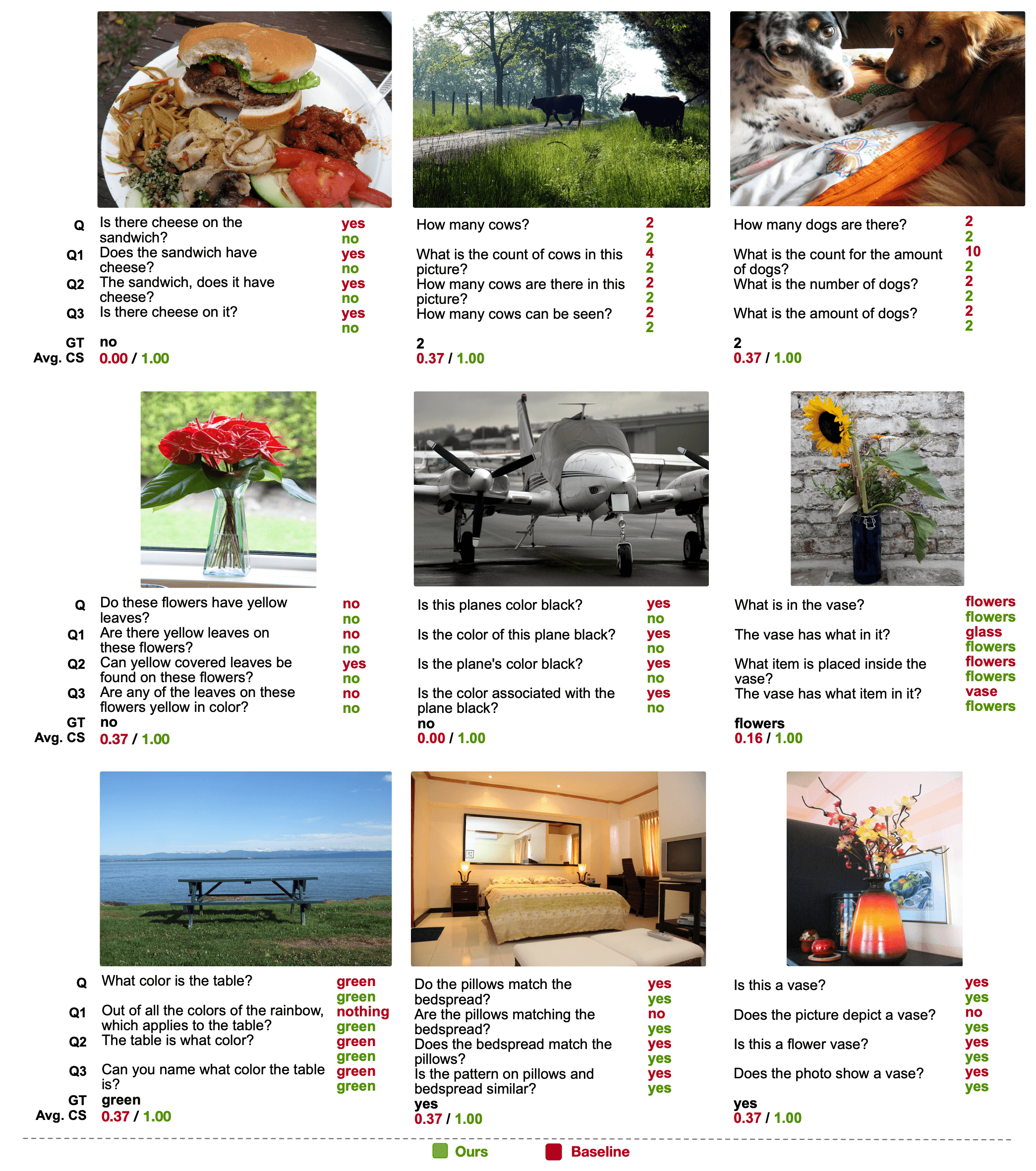}
  \caption{Qualitative Examples. Predictions of \concat/ and MMT+CE baseline on several image-question pairs and their corresponding rephrased questions. Average Consensus Scores (k=1-4) are also shown at the bottom (higher the better).}
  \label{fig:qualitative2}
\end{figure*}

\begin{figure*}[t!]
  \centering
    \renewcommand{\thefigure}{D}
  \includegraphics[width=0.9\textwidth]{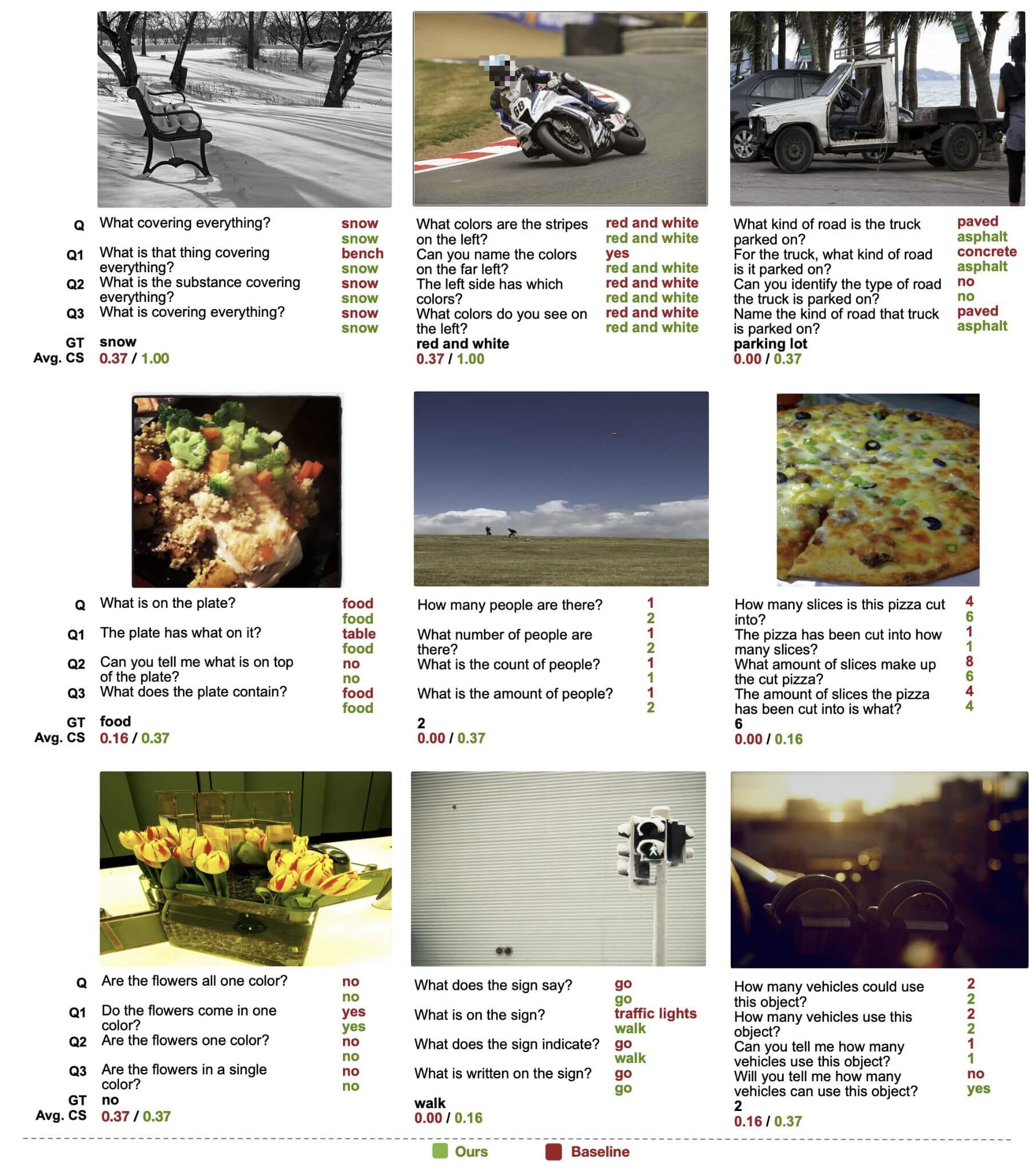}
  \caption{Qualitative Examples. Predictions of \concat/ and MMT+CE baseline on several image-question pairs and their corresponding rephrased questions. Average Consensus Scores (k=1-4) are also shown at the bottom (higher the better).}
  \label{fig:qualitative3}
\end{figure*}

\begin{figure*}[t!]
  \centering
    \renewcommand{\thefigure}{E}
  \includegraphics[width=0.9\textwidth]{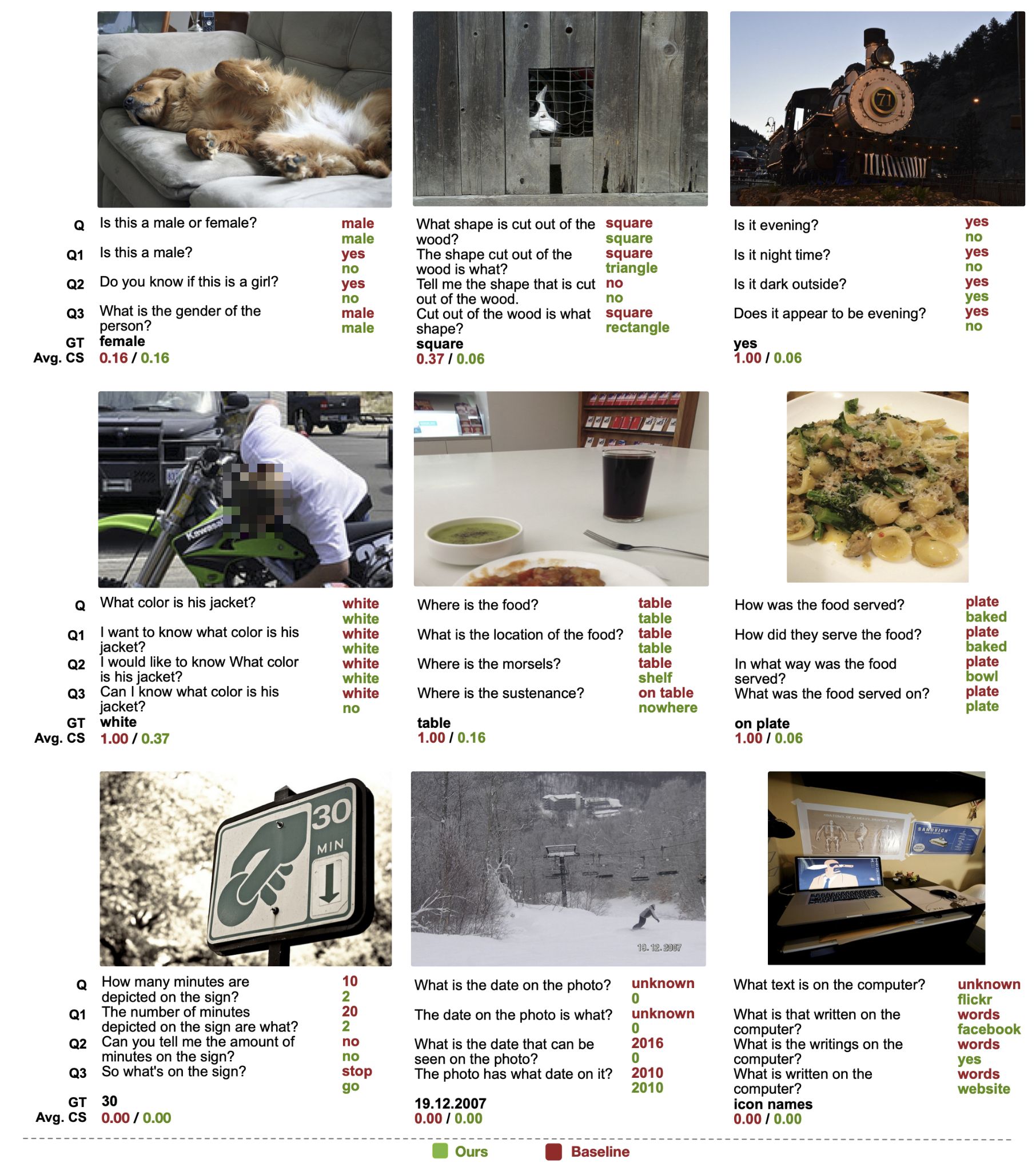}
  \caption{Qualitative Examples. Predictions of \concat/ and MMT+CE baseline on several image-question pairs and their corresponding rephrased questions. Average Consensus Scores (k=1-4) are also shown at the bottom (higher the better).}
  \label{fig:qualitative4}
\end{figure*}

\begin{figure*}[t!]
  \centering
    \renewcommand{\thefigure}{F}
  \includegraphics[width=0.9\textwidth]{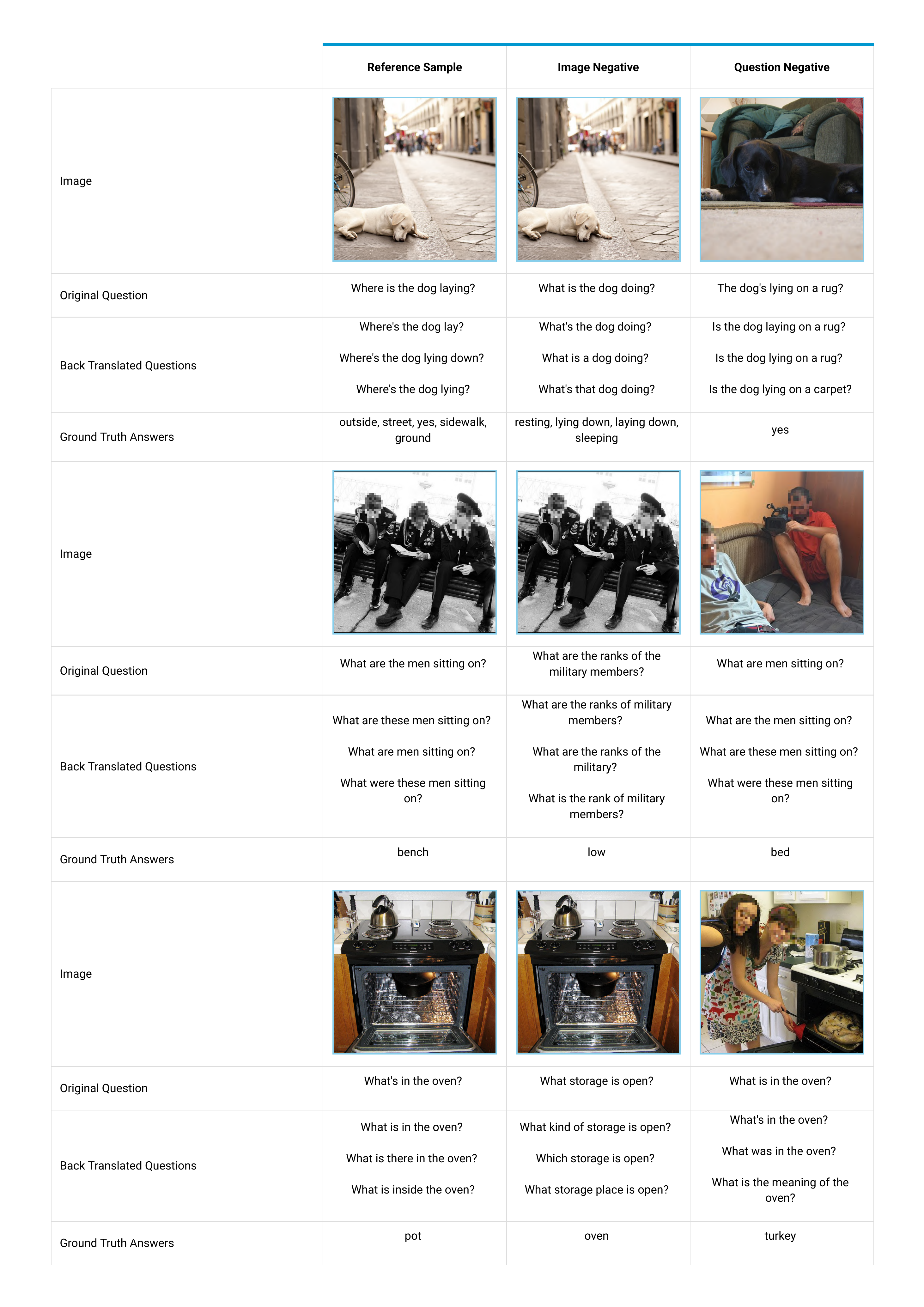}
  \caption{Visualizing the triplets of samples from VQA dataset with corresponding mined Image and Question Negatives.}
  \label{fig:qualitative_bt1}
\end{figure*}

\begin{figure*}[t!]
  \centering
    \renewcommand{\thefigure}{G}
  \includegraphics[width=0.9\textwidth]{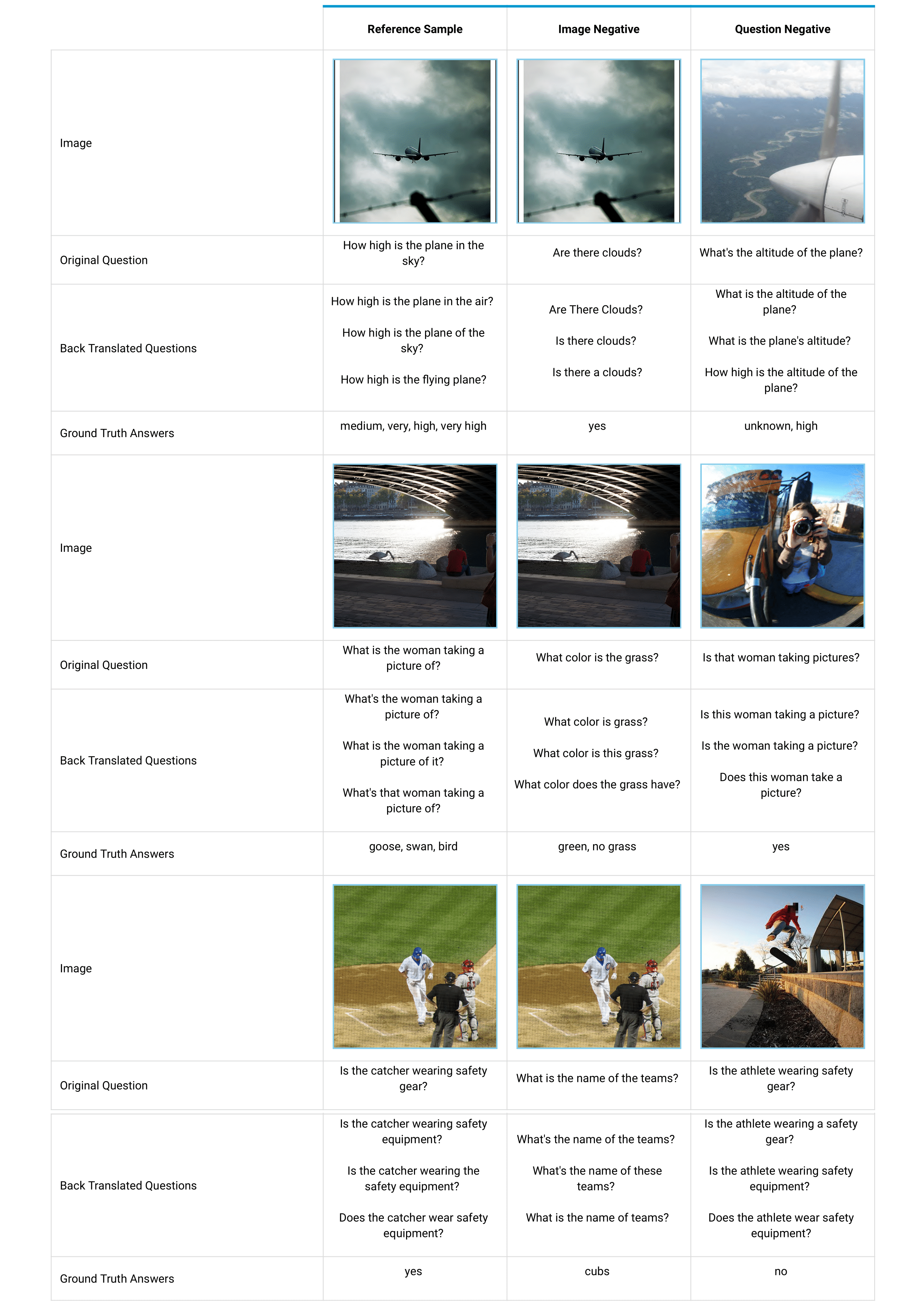}
  \caption{Visualizing the triplets of samples from VQA dataset with corresponding mined Image and Question Negatives.}
  \label{fig:qualitative_bt2}
\end{figure*}

\begin{figure*}[t!]
  \centering
    \renewcommand{\thefigure}{H}
  \includegraphics[width=0.9\textwidth]{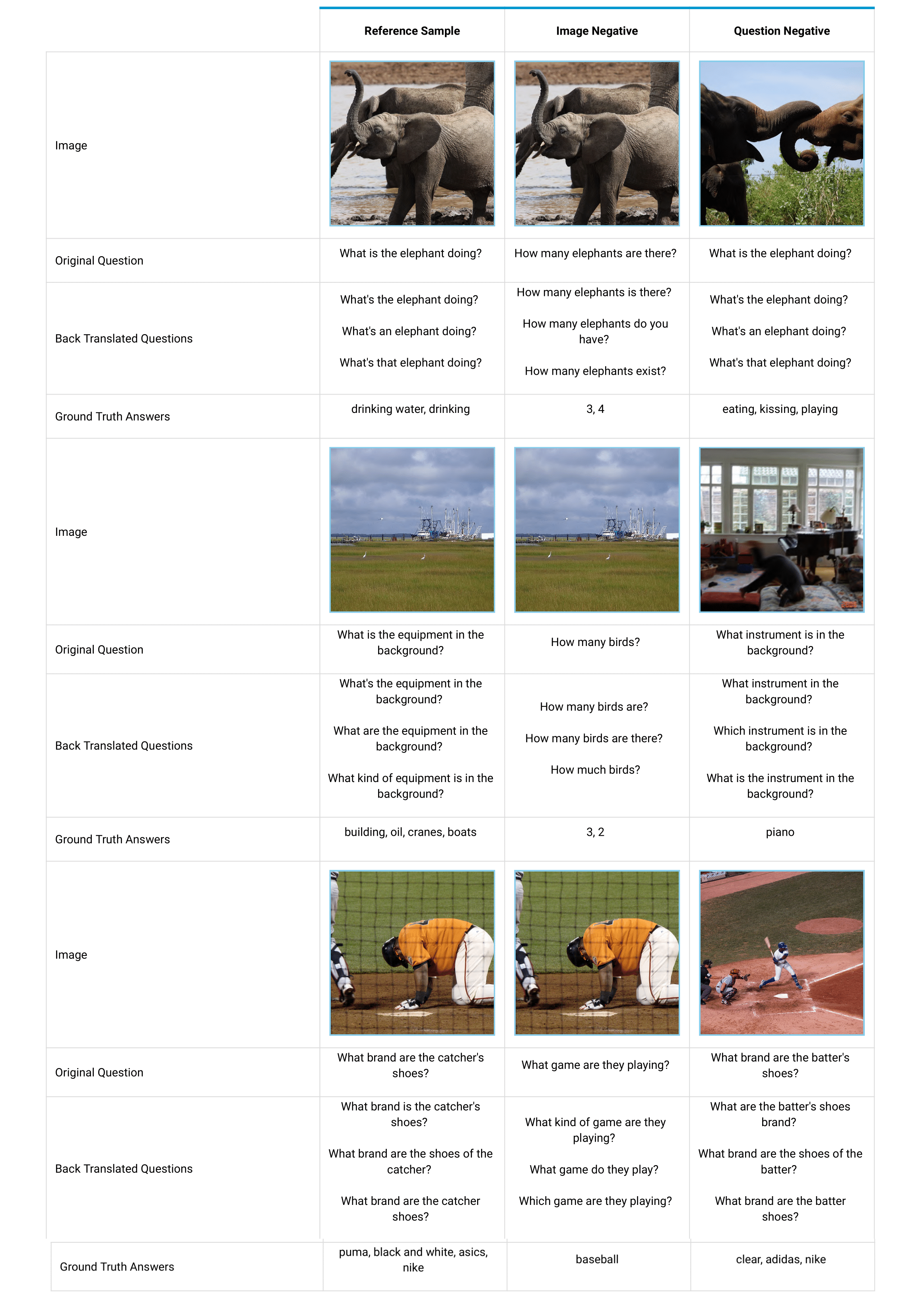}
  \caption{Visualizing the triplets of samples from VQA dataset with corresponding mined Image and Question Negatives.}
  \label{fig:qualitative_bt3}
\end{figure*}

Figures~\ref{fig:qualitative1}, \ref{fig:qualitative2}, \ref{fig:qualitative3}, \ref{fig:qualitative4} show many more qualitative samples comparing the baseline and \concat/. We visualize the data generated via Back-translation and mined triplets in Figures~\ref{fig:qualitative_bt1}, \ref{fig:qualitative_bt2}, \ref{fig:qualitative_bt3}. 

%% file: tables/joint_ablate.tex
\newcolumntype{H}{>{\setbox0=\hbox\bgroup}c<{\egroup}@{}}
\begin{table}[h!]\centering
\setlength{\tabcolsep}{3.5pt}
\renewcommand{\thetable}{A}
\begin{tabular}{clHcHHHHccHHHHH}
 \toprule
  \multirow{2}{*}{} & \multirow{2}{*}{\textbf{Model}} & \multirow{2}{*}{\textbf{DA}} & \multirow{2}{*}{\textbf{$\beta$}}& \multirow{2}{*}{\textbf{$\beta$}} & \multicolumn{4}{c}{\multirow{2}{*}{\textbf{CS(4)}}} & \multicolumn{1}{c}{\textbf{VQA}} \\
  ~ &  ~ &  ~ & ~ & ~ & \textbf{k=1} & \textbf{k=2} & \textbf{k=3} &  & \textbf{val} & \textbf{test-dev}&  \textbf{test-std} \\
 \midrule
 \band \small\texttt{1} & MMT   & BT & 0.25 & - & 67.86 & 60.50 & 56.07 & {52.97} & 66.14 &  &  \\
 \small\texttt{2} & MMT   & BT & 0.50 & - & 67.58 & 60.04 & 55.53 & 53.63 & {66.23} & &\\
 \band \small\texttt{3} & MMT   & BT & 0.75 & - & 63.55 & 56.13 & 51.68 & 48.53 & 61.34 & - & - \\
 \small\texttt{4} & MMT   & BT & 0.90 & - & 54.20 & 47.33 & 43.40 & 40.68 & 51.03 & - & - \\
% \band  \small\texttt{7} & MMT  & CE + SCL (J-R) & BT & \xmark & \texttt{R} & 67.22 & 59.92 & 55.54 & 52.43 & 65.57 & 67.13 & -1 \\
%  \small\texttt{7} & MMT  &  + SCL (J-S) & BT & \xmark & \texttt{R} & 66.95 & 59.70 & 55.32 & 52.20 & 65.10 & 66.73 & -1 \\
\band \small\texttt{5} & MMT + \concat/   & BT & - & \texttt{RQI} & \textbf{68.62} & \textbf{61.42} & \textbf{57.08} & \textbf{53.99} & \textbf{66.98} & {69.80} & \textbf{70.00} \\

 \bottomrule

\end{tabular}
  \caption{Ablations on the choice of our hyper-parameter $\beta$ for joint training.}
  \label{joint-ablation}
\end{table}

%% file: tables/supp_hyper.tex
\setcounter{table}{0}
\renewcommand{\thetable}{B}
\begin{table*}[h!]\scriptsize
\setlength\tabcolsep{3.5pt}
\renewcommand{\arraystretch}{1.25}
\center
\caption{Hyperparameter choices for models.} 
\resizebox{0.8\textwidth}{!}{
\begin{tabular}{clccclc}
\toprule
\# & Hyperparameters & Value & & \# & Hyperparameters & Value\\
\midrule
\band \small\texttt{1} & Maximum question tokens & 23 &&
\small\texttt{2} & Maximum object tokens & 101 \\
\small\texttt{3} & \lce/:\lsscl/ iterations ratio & 3:1 &&
\small\texttt{4} & Number of TextBert layers & 3 \\
\band \small\texttt{5} & Embedding size & 768 &&
\small\texttt{6} & Number of Multimodal layers & 6 \\
\small\texttt{7} & Multimodal layer intermediate size & 3072  &&
\small\texttt{8} & Number of attention heads & 12  \\
\band \small\texttt{9} & Negative type weights ($\bm w$) &  $(0.25, 0.25, 0.5)$ &&
\small\texttt{10} & Multimodal layer dropout & 0.1 \\
\small\texttt{11} & Similarity Threshold ($\epsilon$) & 0.95 &&
\small\texttt{12} & Optimizer & Adam \\
\band \small\texttt{13} & Batch size & 210/420 &&
\small\texttt{14} & Base Learning rate & 2e-4 \\
\small\texttt{15} & Warm-up learning rate factor & 0.1 &&
\small\texttt{16} & Warm-up iterations & 4266 \\
\band \small\texttt{17} & Vocabulary size & 3129 &&
\small\texttt{18} & Gradient clipping (L-2 Norm) & 0.25 \\
\small\texttt{19} & Number of epochs & 5/20 &&
\small\texttt{20} & Learning rate decay & 0.2 \\
\band \small\texttt{21} & Learning rate decay steps & 10665, 14931 &&
\small\texttt{22} & Number of iterations & 25000 \\
\band \small\texttt{23} & Projection Dimension ($\mathcal{R}^{d_z}$) & 128 &&
\small\texttt{24} & Scaling Factor ($s$) & 20 \\
\small\texttt{25} & $N_{ce}$ & 4 &&
\small\texttt{26} & $N_{r}$ & 70 \\

\bottomrule
\end{tabular}
}
\label{tab:hyperparameters_supp}
\end{table*}

%% file: tables/iccv_ablations_full.tex
\newcolumntype{H}{>{\setbox0=\hbox\bgroup}c<{\egroup}@{}}
\begin{table*}[h!]\centering
\setlength{\tabcolsep}{3.5pt}
\begin{tabular}{clcccccccccHH}
 \toprule
  \multirow{2}{*}{} & \multirow{2}{*}{\textbf{Model}} & \multirow{2}{*}{\textbf{Loss(es)}} & \multirow{2}{*}{\textbf{Scaling}}& \multirow{2}{*}{\textbf{N-Type}} & \multirow{2}{*}{\textbf{Train Scheme}} & \multirow{2}{*}{\textbf{CS(1)}}  & \multirow{2}{*}{\textbf{CS(2)}} & \multirow{2}{*}{\textbf{CS(3)}} & \multirow{2}{*}{\textbf{CS(4)}} & \multicolumn{3}{c}{\textbf{VQA}} \\

  ~ &  ~ &  ~ & ~ & ~ & \textbf{~} & \textbf{~} & ~ & ~ & ~ & \textbf{val} & \textbf{test-dev}&  \textbf{test-std} \\
 \midrule
 \small\texttt{1} & MMT   & \lce{} & - & - & - &67.58 & 60.04 & 55.53 & 52.36 & 66.31 & 69.51 & 69.22 \\
  \small\texttt{2} & MMT   & \lsscl{} \& \lce{} & \ding{51} & \texttt{R} & Alternate & 68.19 & 60.92 & 56.53 & 53.42 & 66.62 & - & -\\
\band \small\texttt{3} & MMT  & \lscl{} \& \lce{} & \ding{51} & \texttt{RQ} & Alternate & 68.41 & 61.24 & 56.88 & 53.77 & 66.97 & - & -\\
 \small\texttt{4} &  MMT   & \lsscl{} \& \lce{} & \ding{51} & \texttt{RI} & Alternate & 68.47 & 61.28 & 56.91 & 53.79 & 66.93 & - & -\\
\small\texttt{5} & MMT   & \lsscl{} \& \lce{} & \ding{51} & \texttt{QI} & {Alternate} & {68.65} & {61.40} & {57.00} & {53.90} & {66.95} & - & - \\
\small\texttt{6} & MMT   & \lsscl{} \& \lce{} & \ding{51} & \texttt{RQI} & {Alternate} & \textbf{68.62} & \textbf{61.42} & \textbf{57.08} & \textbf{53.99} & \textbf{66.98} & {69.80} & \textbf{70.00} \\

\midrule
\band \small\texttt{7} & MMT   & \lscl{} \& \lce{} & \xmark & \texttt{RQI} & Alternate & 68.20 & 60.90 & 56.49 & 53.36 & 66.60 & - & -\\

\small\texttt{8} & MMT & \lsscl{} \& \lce{} & Dynamic & \texttt{RQI} & Alternate & 68.60 & 61.38 & 57.01 & 53.92 & 66.95 & - & - \\

\midrule
\band \small\texttt{9} & MMT  & \lsscl{} \& \lce{} & \ding{51} & \texttt{RQI} & Joint & 67.75 & 60.79 & 56.59 & 53.63 & 66.23 & - & - \\

 \small\texttt{10} & MMT   & \lsscl{} $\rightarrow$ \lce{}~\cite{Khosla2020SupervisedCL} & \xmark & \texttt{RQI}& Pretrain-Finetune & 65.33 & 57.39 & 52.63 & 49.20 & 64.21 & - & - \\

\midrule

 \band \small\texttt{11} & MMT  & $\mathcal{L}_{\text{DMT}}$~\cite{free_vqa} \& \lce{} & \xmark & \texttt{RQI}& Alternate & 68.11 & 60.70 & 56.23 & 53.10 & 66.59 & 68.28 & 68.38 \\  

\bottomrule

\end{tabular}
  \caption{\textbf{Ablations Study of \concat/}. \textbf{Scaling} denotes whether scaling factor $\alpha$ was used. \textbf{N-Type} defines the type of negatives used from Image (\texttt{I}), Question (\texttt{Q}) and Random (\texttt{R}). All experiments are run with Back Translation data.}
  \label{ablation_table_supp}
\end{table*}